\documentclass{article}
\usepackage[T1]{fontenc}     
\usepackage{ragged2e}        
\usepackage[protrusion=true,expansion=false]{microtype}  
\usepackage{graphicx}
\usepackage{multirow}
\usepackage{caption}
\usepackage{float}
\usepackage{booktabs}
\usepackage{multicol}
\usepackage{tablefootnote}
\usepackage{siunitx}
\usepackage{tabularx}        
\usepackage[table,xcdraw]{xcolor}
\usepackage{amsmath,amssymb,amsfonts}
\usepackage{mathtools} 
\usepackage{amsthm}
\usepackage{mathrsfs}
\usepackage{algorithm}
\usepackage{algpseudocode}  
\usepackage{algorithmicx}

\usepackage{textcomp}
\usepackage{pifont}
\usepackage{setspace,lineno}
\usepackage{listings}
\usepackage{soul}            
\sethlcolor{yellow}          
\usepackage[title]{appendix}
\usepackage{cite}
\usepackage{manyfoot}
\usepackage{xcolor}
\definecolor{customgreen}{HTML}{00B050}
\definecolor{captionblue}{HTML}{0070C0}  

\DeclareCaptionLabelFormat{boldblue}{\textbf{\textcolor{captionblue}{#1 #2}}}
\usepackage{titlesec}
\titleformat{\section}
  {\normalfont\Large\bfseries\color{captionblue}}{\thesection}{1em}{}
\titleformat{\subsection}
  {\normalfont\large\bfseries\color{captionblue}}{\thesubsection}{1em}{}
\titleformat{\subsubsection}
  {\normalfont\normalsize\bfseries\color{captionblue}}{\thesubsubsection}{1em}{}
\usepackage{geometry}
\usepackage{amsmath} 
\allowdisplaybreaks[4]
\usepackage[protrusion=true,expansion=false]{microtype}
\usepackage{hyperref}
\hypersetup{
    colorlinks=true,
    linkcolor=blue,
    citecolor=blue,
    urlcolor=blue
}
\title{CLAIRE: A Dual Encoder Network with RIFT Loss and Phi-3 Small Language Model Based Interpretability for Cross-Modality Synthetic Aperture Radar and Optical Land Cover Segmentation}
\author{
Debopom Sutradhar\textsuperscript{1}, 
Arefin Ittesafun Abian\textsuperscript{1}, 
Mohaimenul Azam Khan Raiaan\textsuperscript{1},\\
Reem E. Mohamed\textsuperscript{2}, 
Sheikh Izzal Azid\textsuperscript{3}, 
Sami Azam\textsuperscript{4,*}\\
\small
\textsuperscript{1} Department of Computer Science and Engineering, United International University, Dhaka 1212, Bangladesh\\
\small
\textsuperscript{2} Faculty of Science and Information Technology, Charles Darwin University, Sydney, NSW, Australia 
 \\
\small
\textsuperscript{3} School of Engineering and Energy , Murdoch University , Murdoch , WA 6150, Australia \\
\small
\textsuperscript{4} Faculty of Science and Technology, Charles Darwin University, Casuarina, NT 0909, Australia\\
\small 
\textsuperscript{*} Corresponding Author: sami.azam@cdu.edu.au
}
\date{} 
\begin{document}
\justifying
\twocolumn[
\maketitle
\begin{abstract} 
\noindent
Accurate land cover classification from satellite imagery is crucial in environmental monitoring and sustainable resource management. However, it remains challenging due to the complexity of natural landscapes, the visual similarity between classes, and the significant class imbalance in the available datasets. To address these issues, we propose a dual encoder architecture that independently extracts modality-specific features from optical and Synthetic Aperture Radar (SAR) imagery, which are then fused using a cross-modality attention-fusion module named Cross-modality Land cover segmentation with Attention and Imbalance-
aware Reasoning-Enhanced Explanations (CLAIRE). This fusion mechanism highlights complementary spatial and textural features, enabling the network to better capture detailed and diverse land cover patterns. We incorporate a hybrid loss function that utilizes Weighted Focal Loss and Tversky Loss named RIFT (Rare-Instance Focal-Tversky) to address class imbalance and improve segmentation performance across underrepresented categories. Our model achieves competitive performance across multiple benchmarks: a mean Intersection over Union (mIoU) of 56.02\% and Overall Accuracy (OA) of 84.56\% on the WHU-OPT-SAR dataset; strong generalization with a mIoU of 59.89\% and OA of 73.92\% on the OpenEarthMap-SAR dataset; and remarkable robustness under cloud-obstructed conditions, achieving an mIoU of 86.86\% and OA of 94.58\% on the PIE-RGB-SAR dataset. Additionally, we introduce a metric-driven reasoning module generated by a Small Language Model (Phi-3), which generates expert-level, sample-specific justifications for model predictions, thereby enhancing transparency and interpretability. 

\end{abstract}

\vspace{0.5em}
\vspace{1em}
]

\section{Introduction}

As global environmental challenges persist, the need for precise mapping and monitoring of the Earth's surface has become increasingly essential \cite{yue2024bclnet}. Land cover classification (LUC) employing remote sensing imagery is important to support urban planning, environmental protection, disaster response, and geospatial analysis \cite{zhang2025flexisam}. However, accurate land use classification remains challenging due to spectral similarity between classes, seasonal variations, and the difficulty of effectively integrating multimodal data such as optical and SAR imagery \cite{li2024remote, liu2024multimodal, vivone2024deep}. Recent advances in satellite technology \cite{rogers2024sea, xu2025diffsarshipinst, paprocki2025estimation, zhou2024gaussian, hashemi2024review} have enabled the acquisition of diverse data, such as optical images and synthetic aperture radar (SAR) data, from multiple sensors, thus improving the accuracy and depth of the analyses \cite{yue2024bclnet}. 

Optical images provide detailed textures and rich spectral information, rendering them valuable for the identification of surface features. However, they rely on sunlight and are frequently hampered by cloud cover and weather conditions \cite{liu2023joint} \cite{mao2025application}. In contrast, SAR effectively captures structural details regardless of lighting or weather conditions, thus constituting a reliable alternative \cite{liu2023joint}. By merging data from optical and SAR sources, the individual strengths of each modality can be used to generate more comprehensive and accurate land cover maps \cite{liu2025oshfnet}. Moreover, the emergence of deep learning and computer vision techniques has further advanced this field by enabling the development of automated segmentation systems, significantly improving the efficiency and precision of land cover mapping using fused remote sensing data \cite{yue2024bclnet}.

Various multimodal fusion strategies have been explored to improve land cover classification using optical and SAR imagery \cite{li2025multimodal, tu2025cloud, zhao2025text}. In recent developments, the focus has shifted towards cross-attention \cite{li2025tacmt} and global fusion \cite{li2025glcd} mechanisms, facilitating the more effective alignment of multiscale features and enhancing contextual comprehension across various modalities. Models such as SwinTFNet \cite {ren2024swintfnet}, MFFNet  \cite{wang2024mffnet}, and JoiTriNet \cite{liu2023joint} demonstrate that the integration of attention modules and the encoding-decoding level fusion can contribute to noticeable improvements in classification accuracy. However, these methods often rely heavily on global context while overlooking fine-grained details, limiting their ability to represent heterogeneous landscapes. 
Local attention and feature-fusion networks, including DEN \cite{gao2024new}, OPT-SAR-MS2Net \cite {hu2024opt}, and BCLNet  \cite{yue2024bclnet} attempt to capture detailed modality-specific features. While these approaches achieve improvements in segmentation boundaries and fine-structure recognition, they typically lack a unified mechanism to balance global and local representations. Moreover, many such networks fail to handle the significant class imbalance present in real-world datasets, which disproportionately reduces performance on underrepresented land cover types.

Although multimodal land cover segmentation techniques have made significant progress, they still have several limitations. Numerous studies focus on local or global fusion, often without an integrated approach to capture contextual and fine-grained features \cite{wang2024mffnet}, \cite{liu2023joint}. Incorrect alignment between optical and SAR modalities and cross-modal inconsistency remain persistent issues. Furthermore, class imbalance is commonly overlooked, leading to degraded performance in minority classes \cite{quan2024learning}, \cite{liu2023joint}. Although attention mechanisms improve performance, they often do not fully utilize complementary cross-modal features \cite {ren2024swintfnet},  \cite {liu2025oshfnet}. These gaps restrict their reliability in real-world applications. Despite these advances, persistent challenges remain such as  misalignment and inconsistency between optical and SAR modalities, limited exploitation of complementary features across modalities,  neglect of class imbalance in training, and  lack of reasoning or explanation mechanisms for model predictions. Current methods prioritize accuracy but provide little transparency, which restricts their adoption in operational decision-making contexts. To our best knowledge, no studies have explored post hoc reasoning mechanisms, which limits the transparency and interpretability of predictions.

To address this gap, CLAIRE introduces a metric-driven reasoning component that generates expert-level, sample-specific explanations for each prediction. This reasoning analysis is valuable because it moves beyond raw performance metrics, offering transparency and actionable insights into why a prediction was made, which classes remain uncertain, and how cross-modal contributions influenced the outcome. Such interpretability not only strengthens user trust but also facilitates model refinement and deployment in high-stakes environmental monitoring scenarios.

Our method achieved a mean Intersection-over-Union (mIoU) of 56.02\% and an overall accuracy (OA) of 84.56\% on the WHU-OPT-SAR dataset, surpassing the performance of existing models. On the OpenEarthMap-SAR dataset, which presents a significant class imbalance, the model maintained strong generalization, reaching an mIoU of 59.89\%, and OA of 73.92\%. Particularly under adverse conditions such as cloud cover in the PIE-RGB-SAR dataset. The proposed cross-modality fusion framework demonstrated remarkable robustness, attaining an OA of 94.58\% and an mIoU of 86.86\%. Furthermore, we introduce a metric-driven reasoning component using a Small Language Model (Phi-3), enabling the generation of expert-level, sample-specific justifications for each prediction. This addition significantly improves transparency and to the best of our knowledge, this is the first attempt to combine advanced multimodal fusion with post-hoc reasoning in land cover segmentation, offering a comprehensive solution that improves both predictive accuracy and interpretability.

The main contributions of this work are listed below. 
\begin{itemize}
    \item Introduces CLAIRE, a land cover segmentation framework that integrates the CMAF module to combine multi-scale and cross-modal features. CMAF employs channel-wise attention, spatial attention, and gating mechanisms to selectively emphasize significant information from optical and SAR modalities. This advanced fusion strategy enhances effectiveness of CLAIRE in handling complex land cover segmentation tasks across diverse input sources.
    \item Designs RIFT, a combined loss function by incorporating Weighted Focal Loss and Tversky Loss to minimize class imbalance. This loss design drives the algorithm to focus more on minority categories by imposing higher penalties for misclassifying underrepresented classes. As a result, it increases the model's robustness and assures more balanced land cover segmentation performance.
    \item Integrates cross-modal fusion and gating effectively combine optical and SAR properties to overcome their respective limitations. While optical images suffer from cloud cover and SAR lacks significant detail, their combination provides beneficial qualities, which improves land cover classification even in cloud-affected areas.
    \item Establishes a metric-guided explainability framework based on the small language model (SLM) to improve the interpretability of our multimodal segmentation outputs. With the help of quantitative insights obtained from the internal outputs of the model, this process offers concise, transparent explanation for every prediction.
    
\end{itemize}

The remainder of this paper is structured as follows. Section \ref{rw} reviews the related work in the field, while Section \ref{method} outlines the methodology of the proposed model. Section \ref {es} describes the experimental setup, including the datasets used and evaluation metrics. Section \ref {result} presents the results along with detailed analysis, and Section \ref {reasonning} introduces the language-based reasoning framework. Section \ref{discussion} provides a comprehensive discussion, and Section \ref {conclusion} concludes the study by summarizing the key findings.

\section{Related Work}
\label{rw}
In this section, we review recent studies on land cover classification, with a particular emphasis on cross-attention mechanisms, global fusion, local attention, and feature fusion strategies. These techniques have recently emerged as key approaches to improving the accuracy of land classification tasks.
\subsection{Cross-Attention and Global Fusion-Based Networks}
Cross-attention and global fusion-based networks have shown superior results in land cover classification by facilitating more efficient alignment of multiscale features and capturing global contextual relationships, as highlighted in recent work \cite{liu2023joint, ren2024swintfnet, wang2024mffnet}. Ren et al. \cite{ren2024swintfnet} introduced SwinTFNet, a dual-stream fusion network utilizing Transformers for global context analysis and incorporates a Cross-Attention Fusion Module to incorporate optical and SAR characteristics. In the XI'AN and POHANG datasets, the model achieved mIoU scores of 76.36 and 76.18, OA of 89.65\% and 90.57\%, and Kappa coefficients of 0.85 and 0.87, respectively. In another paper, Wang et al. \cite{wang2024mffnet} proposed MFFNet, a dual-stream land cover classification network that combines ResNet's deep optical features and PidiNet's SAR edge features. An updated Attention Feature Fusion module was provided to enable successful multimodal fusion across both low- and high-level features, and Atrous Spatial Pyramid Pooling was used to improve global contextual dependencies. This method achieved 81.29\% OA, 50.64\% mIoU, and 73.59\% Kappa on the WHU-OPT-SAR dataset. Similarly, Liu et al. \cite{liu2023joint} proposed a joint network featuring encoding-level fusion (JoiTriNet-e) and decoding-level fusion (JoiTriNet-d) to integrate optical and SAR modalities for land cover classification. The network incorporates a multimodal dual-attention fusion module to combine heterogeneous features. JoiTriNet-e achieved 85.73\% OA, 82.63\% Kappa, 76.44\% mPA, and 65.77\% mIoU, and JoiTriNet-d performed slightly better than 86.06\% OA, 83.04\% Kappa, 77.19\% mPA, and 66.58\% mIoU in the DFC2020 dataset.

\subsection{Local Attention and Multimodal Feature-Fusion Networks}
By efficiently capturing fine-grained details and combining relevant features from multiple sources, local attention and feature fusion networks optimize the classification of land cover. Recent research \cite{gao2024new,  xu2024multi, quan2024learning, hu2024opt, liu2025oshfnet, yue2024bclnet, zhang2024asanet} has greatly supported the use of these approaches.
To begin with, Gao et al. \cite{gao2024new} constructed SAMFNet, a fusion network termed DEN, which employs two encoders to effectively utilize modality-specific characteristics while keeping complexity low. Another Detail Attention Module is additionally employed to extract tiny details that are apparent in SAR imagery but concealed in optical images. It obtained 77.0\% OA, 30.0\% mIoU, and 50.0\% kappa on the WHU-OPT-SAR dataset. Furthermore, Xu et al.\cite{xu2024multi} proposed a two-branch segmentation framework with a symmetric attention module for irregular object perception and a multiscale fusion module to address cross-modal semantic inconsistency. Using the WHU-OPT-SAR dataset, it obtained 83.34\% OA, 76.52\% kappa, and 50.49\% mIoU. In another paper, for feature classification tasks, Quan et al. \cite{quan2024learning} presented a general multimodal fusion technique (GMFNet) that can be integrated with most encoding-decoding structures. In the WHU-OPT-SAR dataset, it obtained an accuracy of 81.16\%, 64.98\% mPA, and 53.37\% mIoU.

\begin{table*}[ht!]
\centering
\caption{Performance of state-of-the-art multimodal land cover segmentation models}
\label{tab:comparison}
\scriptsize
\begin{tabular}{|>{\raggedright\arraybackslash}p{2.2cm}|>{\centering\arraybackslash}p{1.3cm}|>{\centering\arraybackslash}p{1.3cm}|>{\centering\arraybackslash}p{1.8cm}|>{\centering\arraybackslash}p{0.9cm}|>{\centering\arraybackslash}p{0.9cm}|>{\centering\arraybackslash}p{0.9cm}|>{\raggedright\arraybackslash}p{4.0cm}|}
\hline
\textbf{Method} & \textbf{Paper} & \textbf{Model} & \textbf{Dataset} & \textbf{mIoU (\%)} & \textbf{Kappa} & \textbf{OA (\%)} & \textbf{Main Limitation} \\
\hline
\multirow{7}{2.2cm}{Cross-Attention and Global Fusion Based Networks} & Ren et al. \cite{ren2024swintfnet} & SwinTFNet & XIAN & 76.36 & 85.0 & 89.65 & \multirow{2}{4cm}{ Strong global context via Transformers but lacks fine-grained focus; ignores class imbalance.} \\
\cline{2-7}
& Ren et al. \cite{ren2024swintfnet} & SwinTFNet & POHANG & 76.18 & 87.0 & 90.57&  \\
\cline{2-8}
& Wang et al. \cite{wang2024mffnet} & MFFNet & WHU-OPT-SAR & 50.64 & 73.59 & 81.29 & Combines deep features with edges, but limited fine detail capture; overlooks class imbalance. \\
\cline{2-8}
& Liu et al. \cite{liu2023joint} & JoiTriNet-e & DFC2020 & 65.77 & 82.63 & 85.73 & Encoder-level fusion only, missing decoding-stage integration; does not tackle class imbalance. \\
\cline{2-8}
& Liu et al. \cite{liu2023joint} & JoiTriNet-d & DFC2020 & 66.58 & 83.04 & 86.06 & Decoder-level fusion only, lacking early-stage integration; class imbalance not addressed. \\

\hline
\multirow{7}{2.2cm}{Attention and Feature-Fusion Networks} & Gao et al. \cite{gao2024new} & DEN & WHU-OPT-SAR & 30.00 & 50 & 77 & Detail attention module, but very low mIoU indicates poor global context integration; no class imbalance handling. \\
\cline{2-8}
& Hu et al. \cite{hu2024opt} & OPT-SAR-MS2Net & WHU-OPT-SAR & 45.2 & — & 84.3 & Multisource multiscale fusion, yet relatively low mIoU suggests insufficient context synergy; no class imbalance handling. \\
\cline{2-8}
& Quan et al. \cite{quan2024learning} & GMFNet & WHU-OPT-SAR & 53.37 & — & 81.16 & General fusion approach yields moderate accuracy; class imbalance overlooked. \\
\cline{2-8}
& C. Liu et al. \cite{liu2025oshfnet} & OSHFNet & WHU-OPT-SAR & 46.56 & 73.96 & 81.65 & Heterogeneous dual-branch with dynamic gating, but incomplete cross-modal feature integration. \\
\cline{2-8}
& Yue et al. \cite{yue2024bclnet} & BCLNet & WHU-OPT-SAR & 51.0 & — & 83.3 & Boundary contrastive learning improves edges but overall mIoU remains moderate; does not address class imbalance. \\
\cline{2-8}
& Zhang et al. \cite{zhang2024asanet} & ASANet & WHU-OPT-SAR & 56.11 & 76.43 & 84.56 & Asymmetric feature fusion improves performance but class imbalance remains; lacks interpretability. \\
\cline{2-8}
& Zhang et al. \cite{zhang2024asanet} & ASANet & PIE-RGB-SAR & 78.31 & 85.27 & 89.64 & No class imbalance handling and no reasoning/interpretability mechanisms. \\

\hline
\multirow{3}{2.2cm}{Dual Encoder Network with Cross-Modality Attention Fusion}
& \multirow{3}{1.3cm}{\textbf{Proposed Study}} & \multirow{3}{1.3cm}{\textbf{CLAIRE (Ours)}} & OpenEarthMap-SAR & 59.89 & — & 73.92 & \multirow{3}{4cm}{Addresses class imbalance with novel loss and adds post-hoc reasoning; achieves robust performance.} \\
\cline{4-7}
&  & & WHU-OPT-SAR & 56.02 & 76.43 & 84.56 & \\
\cline{4-7}
&  &  & PIE-RGB-SAR & 86.86 & 91.76 & 94.58 & \\

\hline
\end{tabular}
\end{table*}

Similarly, Hu et al. \cite{hu2024opt} presented OPT-SAR-MS2Net, a Siamese semantic segmentation network that combines multisource and multiscale fusion modules to extract and fuse optical and SAR features. Using a variety of sensory fields improves contextual learning. Within the WHU-OPT-SAR dataset, it obtained 45.2\% mIoU and 84.3\% OA. In another work, Liu et al. \cite{liu2025oshfnet} presented OSHFNet, a heterogeneous dual-branch network using CNN for optical and VMamba for SAR to extract complementary features, followed by a global-local dynamic gating fusion module that integrates multiscale and cross-modal information through self-attention and dynamic gating. It achieved 81.65\% OA, 73.96\% kappa, and 46.56\% mIoU on the WHU-OPT-SAR dataset. On the other hand, Yue et al. \cite{yue2024bclnet} developed BCLNet, a boundary contrastive learning network that fuses optical and SAR features through a Gate Attention Fusion module and enhances them using a multibranch spatial channel reconstruction module, improving semantic embedding and classification accuracy. It obtained 51.0\% mIoU and 83.3\% overall accuracy on the WHU-OPT-SAR dataset. Finally, Zhang et al. \cite{zhang2024asanet} proposed ASANet, which addresses multimodal feature utilization by introducing feature-level asymmetry using the Semantic Focusing Module, which weights modality-specific features and is enhanced by the Cascade Fusion Module, integrating channel and spatial features efficiently. It got 76.43 kappa, 84.56\% OA, and 56.11\% mIoU on the WHU-OPT-SAR dataset, and 85.27 kappa, 89.64\% OA, and 78.31\% mIoU on the PIE-RGB-SAR dataset. 

\begin{figure*}[ht!]
    \centering
    \includegraphics[width=0.85 \textwidth]{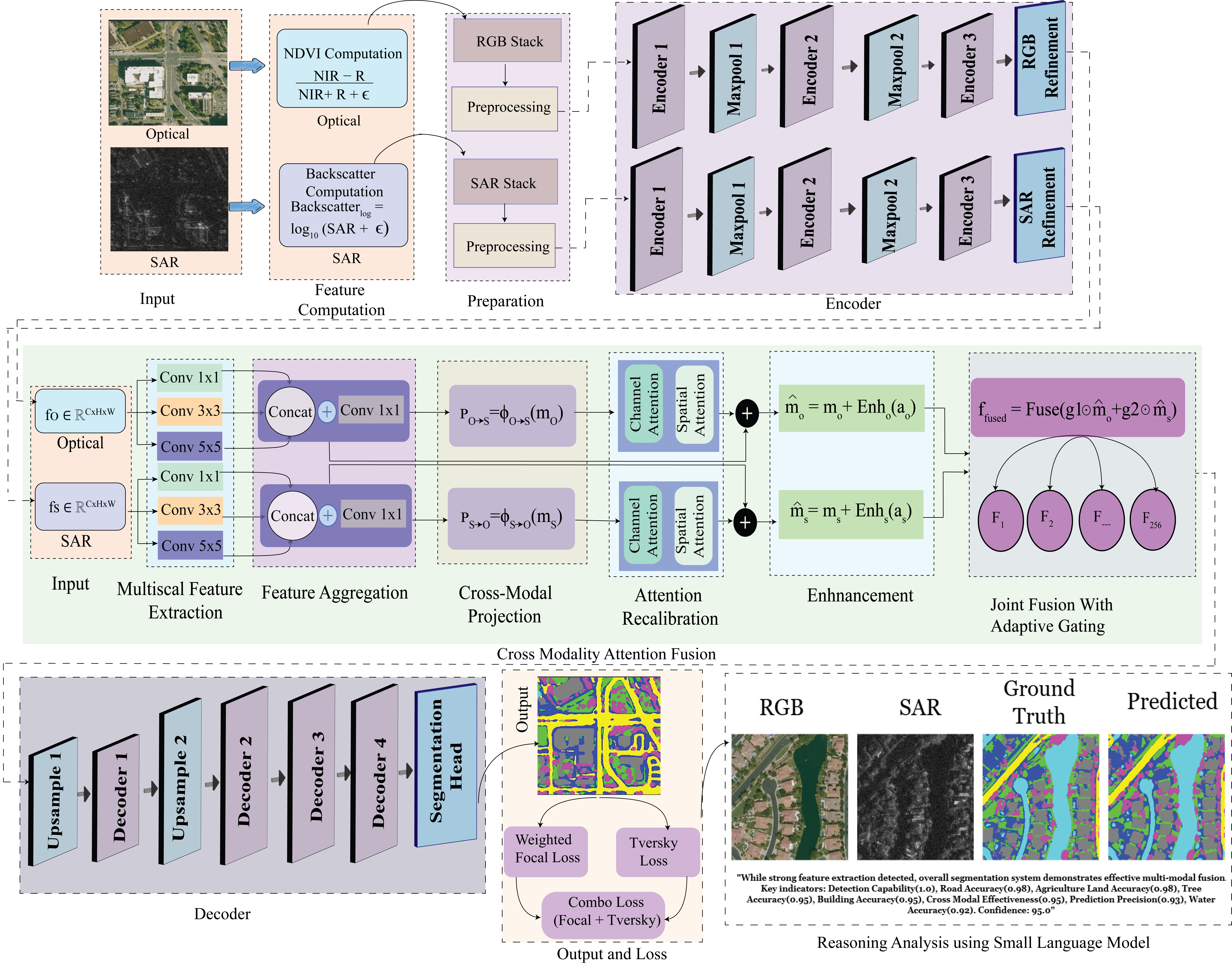} 
    \caption{
        Overview of CLAIRE, our proposed multimodal semantic segmentation framework. The architecture consists of parallel deep encoders for optical and SAR inputs, followed by cross-modality attention fusion and a unified decoder.
    }
    \label{fig:method}
\end{figure*}

Our proposed model, CLAIRE addresses several key limitations observed in existing studies by offering novel insights and solutions, presented in Table \ref{tab:comparison}. Unlike traditional approaches, CLAIRE  effectively handles class imbalance by using two complementary loss functions: Weighted Focal Loss and Tversky Loss. This strategy has not been explored in previous work. Moreover, the CMAF module integrates multiscale and cross-modal features through the use of channel-wise attention, spatial attention, and gating mechanisms to prioritize the most relevant information from both optical and SAR modalities, thereby enhancing its performance in complex land cover segmentation tasks. To the best of our knowledge, this is the first work to introduce a SLM for post hoc reasoning in land class segmentation. This integration empowers the framework not only to generate accurate predictions but also to deliver concise and transparent outputs. In contrast to previous work, which has largely overlooked reasoning-based analysis in segmentation, our approach opens up a new and impactful direction in the field.

While cross-attention networks excel at global feature alignment and feature-fusion approaches enhance fine-grained details, both categories share limitations: they insufficiently address class imbalance, struggle with modality misalignment, and overlook interpretability. None of the reviewed studies incorporates a reasoning mechanism, leaving end users without insights into prediction reliability. These gaps directly motivate the development of CLAIRE, our proposed framework, which introduces novel fusion, loss, and reasoning modules to address these shortcomings. CLAIRE extends beyond prior cross-attention/global fusion models by explicitly addressing class imbalance and introducing a novel reasoning module.

\section{Methodology}
\label{method}
The proposed CLAIRE framework is designed using a dual-branch encoder-decoder architecture, in which optical and SAR images are processed independently through modality-specific deep encoders to extract robust hierarchical features. At the bottleneck, a Cross-Modality Attention Fusion (CMAF) module integrates the information from both branches. The fused features are then decoded through a series of upsampling and convolutional layers to generate dense semantic predictions. Figure \ref{fig:method} provides a comprehensive visual summary of this multimodal semantic segmentation framework.

CLAIRE consists of four key components, each addressing a core challenge in multimodal land cover segmentation. First, a multimodal preprocessing and feature extraction step ensures consistent spatial and spectral alignment between input modalities. Second, the CMAF module integrates multiscale features using channel-wise attention, spatial recalibration, and adaptive gating, allowing the model to emphasize the most informative modality-specific representations. Third, loss optimization techniques (Weighted Focal Loss and Tversky Loss) are employed to mitigate class imbalance and improve learning on underrepresented classes. Finally, a post-hoc reasoning module based on a Small Language Model (SLM) produces sample-specific natural language explanations, enhancing the interpretability and trustworthiness of segmentation outputs.

\subsection{Data Preprocessing}

In this work, we adopt a structured pipeline that ensures spatial alignment, feature enhancement, and normalization across optical and SAR modalities. As input modalities, we utilize two data sources. \textbf{Optical (RGB) imagery} and \textbf{Synthetic Aperture Radar (SAR) intensity images}. All images are divided into fixed-size patches of $256 \times 256$ pixels to facilitate mini-batch training and efficient computation. For each sample, the corresponding patches are extracted from the RGB, SAR, and label images, ensuring strict pixel-level alignment across the modalities and the ground truth.

The pixel values of the labels are clipped to the valid range of class indices, $[0, N-1]$, where $N$ is the total number of classes of land cover. Labels are resized to match the patch size using nearest-neighbor interpolation to preserve the discrete class boundaries.

\subsubsection{SAR Preprocessing}
Synthetic Aperture Radar (SAR) is an active remote sensing modality capable of acquiring high-resolution images regardless of weather conditions or availability in daylight. Unlike optical cameras, SAR systems transmit microwave signals and record the reflected response from the Earth's surface, and later these signals are combined together for imagery \cite{meng2024synthetic}. However, SAR imagery is inherently affected by speckle noise, resulting from the coherent nature of radar signal acquisition. Speckle noise can degrade interpretability and subsequent analysis. Moreover, the dynamic range of SAR intensities is typically wide, often requiring transformation for effective use in deep learning models. A median filter $3 \times 3$ is applied to suppress speckle noise while preserving important structures. and further enhance the discriminative power of the SAR channel by computing its intensity. In addition to the filtered SAR intensity, we compute a logarithmic transformation of the backscatter channel using equation \eqref{eq:1}.

    \begin{equation}
    \label{eq:1}
        \text{Backscatter}_{\log} = \log_{10}(\text{SAR} + \epsilon)
    \end{equation}
    where $\epsilon = 10^{-6}$ prevents numerical instability for zero-valued pixels.

Finally, the SAR input tensor comprises two channels, median-filtered SAR intensity, and log-backscatter.

\subsubsection{Optical Feature Engineering}

Vegetation indices are essential for enhancing the separability of vegetated and non-vegetated land cover classes in remote sensing imagery.

\textbf{For the WHU-OPT-SAR dataset \cite{li2022mcanet},} the optical imagery provides a near-infrared (NIR) band in addition to the standard RGB channels. This allows for the computation of the canonical Normalized Difference Vegetation Index (NDVI) \cite{huang2021commentary}. Equation \eqref{eq:2} represents the mathematical explanation of NDVI.
\begin{equation}
\label{eq:2}
    \text{NDVI} = \frac{\text{NIR} - \text{R}}{\text{NIR} + \text{R} + \epsilon}
\end{equation}
where NIR and R denote the near-infrared and red bands, respectively, and $\epsilon = 10^{-6}$ is used to avoid division by zero. The resulting NDVI map is stacked with the RGB channels, yielding a four-channel optical tensor for model input.

\textbf{For the OpenEarthMap-SAR dataset \cite{xia2023openearthmap},} there was no NIR band. So, we computed the Visible Atmospherically Resistant Index (VARI) \cite{stow2005modis} as a surrogate vegetation indicator, as this dataset lacks the NIR band and provides only RGB imagery. Mathematically, VARI is calculated using equation \eqref{eq:3}
\begin{equation}
\label{eq:3}
    \text{VARI} = \frac{G - R}{G + R - B + \epsilon}
\end{equation}
where G, R, and B are the green, red, and blue channels, respectively. VARI is specifically designed for RGB images and provides enhanced discrimination of green vegetation in the absence of NIR data. The computed VARI map is stacked with the RGB channels, resulting in a four-channel input tensor.

In both cases, all optical channels (including the vegetation index) are linearly normalized to the range $[0, 1]$. Any NaN or infinite values introduced during the calculation are replaced with safe defaults to ensure numerical stability during model training.

This adaptive strategy ensures that the model benefits from the most informative vegetation index available for each dataset, improving the performance of land cover classification under varying sensor constraints. Table \ref{tab: summary} represents the input channels and their description.

\begin{table}[ht!]
\centering
\caption{Summary of input channels after preprocessing}
\label{tab: summary}
\scriptsize
\begin{tabular}{lll}
\hline
\textbf{Input} & \textbf{Channels} & \textbf{Description} \\
\hline
RGB + NDVI & R, G, B, NDVI & Optical features \\
SAR & SAR, log-backscatter & SAR features \\
Label & 1 (integer) & Ground-truth class index \\
\hline
\end{tabular}
\end{table}

This preprocessing pipeline ensures both robustness and consistency, improving the overall performance and reliability of the proposed segmentation model.

\subsection{Model Architecture}

The proposed network is a deep encoder-decoder architecture designed for effective fusion and segmentation of co-registered SAR and optical remote sensing data. The model uses multiscale feature extraction, advanced attention mechanisms, and a cross-modality fusion block to robustly integrate heterogeneous features from both data modalities. 

\subsubsection{Modality-Specific Encoders}

The core of our optical and SAR processing frameworks is the Deep Encoder Block, which is optimized to effectively capture detailed hierarchical representations from multimodal remote sensing data. The goal of each Encoder Block is to progressively transform the input feature maps and enhance their expressiveness while maintaining robust gradient flow and adaptive focus on informative channels.

Each Encoder Block comprises three consecutive convolutional layers. Each layer contains batch normalization and ReLU activation. This layered structure enables the block to capture local spatial dependencies and build more abstract features as information moves deeper into the network. Let $\mathbf{x} \in \mathbb{R}^{C_{in} \times H \times W}$ denote the input tensor to the block.

Deep networks are known to suffer from vanishing gradients, which can hamper effective learning. To mitigate this, we employed a residual connection \cite{he2016deep}. This residual connection is a shortcut path that allows the input to bypass the convolutional stack. If the input and output channels differ, a $1 \times 1$ convolution with batch normalization, denoted as $\text{Proj}(\cdot)$, projects $\mathbf{x}$ to match the dimensionality of $\mathbf{y}$. Equation \eqref{eq:4} gives mathematical explanation of residual block.
\begin{equation}
\label{eq:4}
    \mathbf{r} =
    \begin{cases}
      \mathbf{x}, & \text{if } C_{in} = C_{out} \\
      \text{Proj}(\mathbf{x}), & \text{otherwise}
    \end{cases}
\end{equation}
where $\mathbf{x} \in \mathbb{R}^{C_{in} \times H \times W}$ is the input feature map, $C_{in}$ and $C_{out}$ are the number of input and output channels, respectively, and $\text{Proj}(\cdot)$ is a $1 \times 1$ convolution followed by batch normalization for channel matching. The preliminary output of the block is therefore the sum in element of the transformed features and the shortcut. Then this sum is followed by a ReLU activation function. The process is explained mathematically in equation \eqref{eq:5}.
\begin{equation}
\label{eq:5}
    \mathbf{z} = \text{ReLU}(\mathbf{y} + \mathbf{r})
\end{equation}
where $\mathbf{y} \in \mathbb{R}^{C_{out} \times H \times W}$ is the output of the convolutional stack, and $\text{ReLU}(\cdot)$ denotes the rectified linear unit activation function applied element-wise.

Not all feature channels contribute equally to semantic understanding, especially when integrating heterogeneous data sources such as optical and SAR images \cite{liu2024softformer}. To tackle this, our encoder incorporates a channel attention mechanism based on the Squeeze-and-Excitation (SE) principle \cite{zhang2022sernet}. SE modules offer a highly effective and computationally efficient approach for dynamically recalibrating features channel-wise. This is especially advantageous in deep networks, where it is crucial to concentrate the model capacity on the most relevant information. 

The process begins by summarizing the spatial information of each feature channel into a single representative value through global average pooling. For a given channel $c$, this summary statistic is computed according to equation \eqref{eq:6}.
\begin{equation}
\label{eq:6}
    s_c = \frac{1}{H W} \sum_{i=1}^{H} \sum_{j=1}^{W} z_{c, i, j}
\end{equation}
where $z_{c, i, j}$ denotes the activation in spatial location $(i, j)$ of channel $c$ on the feature map $\mathbf{z} \in \mathbb{R}^{C \times H \times W}$, with $H$ and $W$ as spatial dimensions and $C$ the number of channels.

The resulting vector of channel descriptors $\mathbf{s} = [s_1, \dots, s_C]^\top$ is then transformed by a lightweight two-layer neural network with nonlinear activations. This learns to assign an importance weight to each channel based on global context. The process is explained through equation \eqref{eq:7}
\begin{equation}
\label{eq:7}
    \mathbf{w} = \sigma \left( W_2 \cdot \delta \left( W_1 \cdot \mathbf{s} \right) \right)
\end{equation}
where $W_1 \in \mathbb{R}^{\frac{C}{r} \times C}$ and $W_2 \in \mathbb{R}^{C \times \frac{C}{r}}$ are the learnable weights of the bottleneck layers, $r$ is a reduction ratio (typically $r=16$), $\delta(\cdot)$ is the ReLU function and $\sigma(\cdot)$ denotes the sigmoid function applied element-wise.

These learned weights $\mathbf{w} = [w_1, \dots, w_C]^\top$ are then used to adaptively rescale the original feature map throughout the channel dimension. Equation \eqref{eq:8} represents the mathematical explanation.
\begin{equation}
\label{eq:8}
    \tilde{\mathbf{z}}_c = \mathbf{z}_c \cdot w_c \qquad \forall\ c \in \{1, \ldots, C\}
\end{equation}
where $\tilde{\mathbf{z}}_c$ and $\mathbf{z}_c$ denote the $c$-th channel of the recalibrated and original feature maps, respectively, and $w_c$ is the corresponding channel weight. This channel attention mechanism allows the block to emphasize the most informative features for the task at hand and suppress noise or irrelevant activations, which is especially beneficial for multimodal data, where channel utility can be context-dependent \cite{jin2022delving}.

Thus, the Encoder Block functions as both a robust extractor of spatial features and an adaptive selector of channels. This design facilitates the acquisition of deep and abstract representations while also safeguarding the retention of critical features as the information moves through the network. The integration of convolutional depth, residual connections, and channel attention equips our model to efficiently handle the complex and varied signals present in remote sensing data.

\subsubsection{Cross-Modality Attention Fusion (CMAF)}

To robustly integrate information from both optical and SAR sources, we employ a cross-modeality attention fusion (CMAF) module that operates at the bottleneck of our architecture. Each modality passes through multiscale feature extraction, where the input features are processed in parallel by convolutional layers with different kernel sizes. Specifically, $1 \times 1$, $3 \times 3$, and $5 \times 5$ kernels. The resulting multiscale feature maps are concatenated along the channel dimension and aggregated using a $1 \times 1$ convolution. This produces aggregated modality features $\mathbf{m}_\mathrm{O}$ and $\mathbf{m}_\mathrm{S}$ for the optical and SAR branches. The mathematical explanation is shown in equations~\eqref{eq:9} and~\eqref{eq:10}.
\begin{align}
    \mathbf{m}_\mathrm{O} &= \mathrm{Agg}_\mathrm{O} \left(
        [\mathrm{Conv}_1(\mathbf{f}_\mathrm{O}),\ \mathrm{Conv}_3(\mathbf{f}_\mathrm{O}),\ \mathrm{Conv}_5(\mathbf{f}_\mathrm{O})]
    \right), \label{eq:9} \\
    \mathbf{m}_\mathrm{S} &= \mathrm{Agg}_\mathrm{S} \left(
        [\mathrm{Conv}_1(\mathbf{f}_\mathrm{S}),\ \mathrm{Conv}_3(\mathbf{f}_\mathrm{S}),\ \mathrm{Conv}_5(\mathbf{f}_\mathrm{S})]
    \right). \label{eq:10}
\end{align}
Here, $\mathbf{f}_\mathrm{O} \in \mathbb{R}^{C \times H \times W}$ and $\mathbf{f}_\mathrm{S} \in \mathbb{R}^{C \times H \times W}$ represent the high-level feature maps extracted from the optical and SAR encoder branches.

For cross-modal information transfer, the combined features of each modality are mapped into the latent space of the opposite modality through learnable convolutional transformations, resulting in projected features. $\mathbf{p}_\mathrm{O \to S}$ and $\mathbf{p}_\mathrm{S \to O}$, as given in equations~\eqref{eq:11} and ~\eqref{eq:12}.
\begin{align}
    \mathbf{p}_\mathrm{O \to S} &= \phi_{\mathrm{O \to S}}(\mathbf{m}_\mathrm{O}), \label{eq:11} \\
    \mathbf{p}_\mathrm{S \to O} &= \phi_{\mathrm{S \to O}}(\mathbf{m}_\mathrm{S}). \label{eq:12}
\end{align}
Here, $\phi_{\mathrm{O \to S}}$ and $\phi_{\mathrm{S \to O}}$ denote convolutional mappings parameterized by learnable weights.

Each projected feature map is then reconfigured through both channel and spatial attention mechanisms. Channel attention $\alpha_\mathrm{C}(\cdot)$ and spatial attention $\alpha_\mathrm{S}(\cdot)$ are applied sequentially, and the outputs are multiplied element-wise to generate attention-weighted features, shown in equations~\eqref{eq:13} and~\eqref{eq:14}.
\begin{align}
    \mathbf{a}_\mathrm{O} &= \alpha_\mathrm{C}(\mathbf{p}_\mathrm{S \to O}) \odot \alpha_\mathrm{S}(\mathbf{p}_\mathrm{S \to O}), \label{eq:13} \\
    \mathbf{a}_\mathrm{S} &= \alpha_\mathrm{C}(\mathbf{p}_\mathrm{O \to S}) \odot \alpha_\mathrm{S}(\mathbf{p}_\mathrm{O \to S}). \label{eq:14}
\end{align}

These recalibrated features are further processed by enhancement blocks and combined with the original aggregated features, resulting in modality-adaptive feature maps $\hat{\mathbf{m}}_\mathrm{O}$ and $\hat{\mathbf{m}}_\mathrm{S}$. Mathematically formulated in equations~\eqref{eq:15} and~\eqref{eq:16}.
\begin{align}
    \hat{\mathbf{m}}_\mathrm{O} &= \mathbf{m}_\mathrm{O} + \mathrm{Enh}_\mathrm{O}(\mathbf{a}_\mathrm{O}), \label{eq:15} \\
    \hat{\mathbf{m}}_\mathrm{S} &= \mathbf{m}_\mathrm{S} + \mathrm{Enh}_\mathrm{S}(\mathbf{a}_\mathrm{S}), \label{eq:16}
\end{align}
where $\mathrm{Enh}_\mathrm{O}$ and $\mathrm{Enh}_\mathrm{S}$ are convolutional enhancement modules.

The next stage involves concatenating these recalibrated features along the channel dimension to form a joint representation $\mathbf{f}_\mathrm{cat} = [\hat{\mathbf{m}}_\mathrm{O},\ \hat{\mathbf{m}}_\mathrm{S}]$. This joint feature map is further refined by an additional round of channel and spatial attention, yielding $\mathbf{f}_\mathrm{att} = \alpha_\mathrm{C}(\mathbf{f}_\mathrm{cat}) \odot \alpha_\mathrm{S}(\mathbf{f}_\mathrm{cat}) \odot \mathbf{f}_\mathrm{cat}$. To adaptively balance the contributions of optical and SAR information, a learned gating mechanism is applied. This gating head, performed as a convolutional layer followed by sigmoid activation, produces a set of spatially variable gating masks $\mathbf{g} = \gamma(\mathbf{f}_\mathrm{att}) \in [0, 1]^{2 \times H \times W}$, corresponding to the optical and SAR modalities.

Finally, the recalibrated and gated modality features are fused through a weighted sum, followed by a final fusion transformation consisting of convolutional operations. The final operation is shown in Equation \eqref{eq:17}.
\begin{equation}
\label{eq:17}
    \mathbf{f}_\mathrm{fused} = \mathrm{Fuse} \left( \mathbf{g}_{1} \odot \hat{\mathbf{m}}_\mathrm{O} + \mathbf{g}_{2} \odot \hat{\mathbf{m}}_\mathrm{S} \right),
\end{equation}
where $\mathbf{g}_{1}$ and $\mathbf{g}_{2}$ are the gating masks for optical and SAR branches, respectively, and $\mathrm{Fuse}(\cdot)$ denotes a sequence of convolutions with batch normalization and non-linearity.

\begin{figure*}[ht!]
    \centering
    \includegraphics[width=0.8\textwidth]{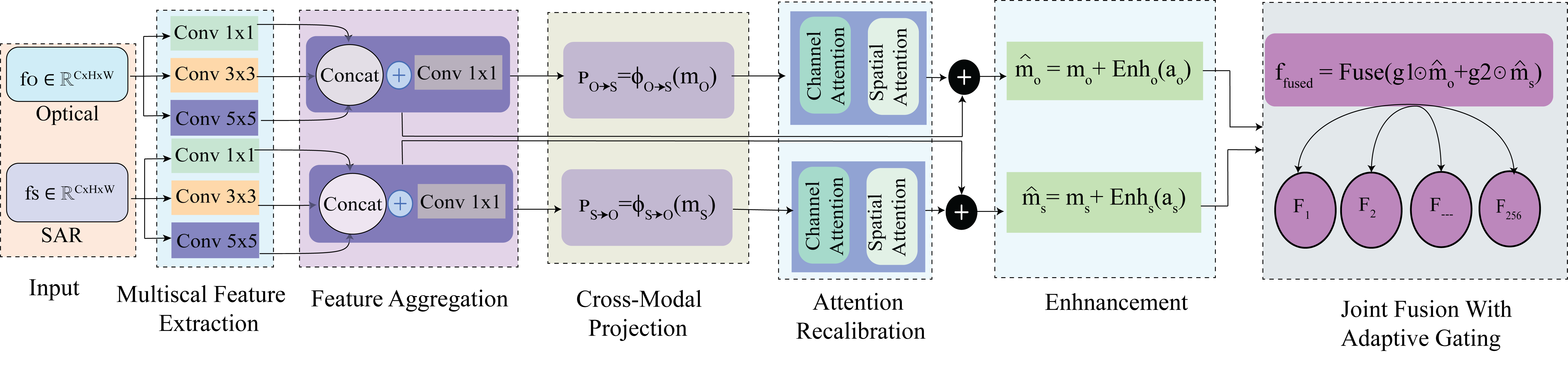} 
    \caption{Visual diagram of Cross-Modality Attention Fusion (CMAF).}
    \label{fig:cmaf}
\end{figure*}

In summary, the CMAF module adaptively integrates multiscale and cross-modality features through channel and spatial attention, as well as learned gating. This helps the network dynamically emphasize the most important and complementary patterns of both modalities. This adaptive fusion significantly enhances the model’s capability to handle challenging land cover segmentation tasks with complex input data. Figure~\ref{fig:cmaf} provides a summarized overview of our CMAF.

\subsubsection{Decoder and Segmentation Head}
The decoder is designed to progressively upsample and refine the fused feature representation, ultimately producing a high-resolution semantic segmentation map. The process begins by upsampling the fused feature map $\mathbf{f}_\mathrm{fused}$ using bilinear interpolation, as shown in Equation~\eqref{eq:18}.
\begin{equation}
    \mathbf{u}_1 = \mathrm{Up}_1(\mathbf{f}_\mathrm{fused}),
    \label{eq:18}
\end{equation}
where $\mathrm{Up}_1(\cdot)$ denotes an upsampling operation with a scale factor of 2. The upsampled feature map is then passed through a sequence of convolutional refinement blocks. Each block is implemented to capture spatial context and further suppress noise. This refinement is performed iteratively in each decoding stage, as expressed in equations~\eqref{eq:19} and~\eqref{eq:20}.
\begin{align}
    \mathbf{d}_1 &= \mathrm{Dec}_1(\mathbf{u}_1), \label{eq:19} \\
    \mathbf{u}_2 &= \mathrm{Up}_2(\mathbf{d}_1), \nonumber \\
    \mathbf{d}_2 &= \mathrm{Dec}_2(\mathbf{u}_2), \label{eq:20}
\end{align}
where $\mathrm{Dec}_k(\cdot)$ represents the $k$-th decoder block (comprising convolution, batch normalization, and ReLU operations), and $\mathrm{Up}_2(\cdot)$ is a second upsampling operation. Further refinement is performed through additional decoder blocks as needed, following the same structure.

After passing through the final decoding block, the network applies a concluding convolutional head to produce the output segmentation logits $\mathbf{y}_\text{pred}$, as shown in Equation~\eqref{eq:21}.
\begin{equation}
    \mathbf{y}_\text{pred} = \mathrm{Head}(\mathbf{d}_\text{final}),
    \label{eq:21}
\end{equation}
where $\mathrm{Head}(\cdot)$ denotes a $3 \times 3$ convolution followed by batch normalization, ReLU activation, and a final $1 \times 1$ convolution to map the feature channels to the number of classes.

In this manner, the decoder effectively restores spatial resolution while preserving and refining the rich information distilled by the encoder and fusion modules. The use of deep convolutional refinement and progressive upsampling enables the model to accurately delineate object boundaries and recover fine-grained spatial details essential for high-quality segmentation.

This architecture allows for expressive and adaptive integration of SAR and optical cues, leading to robust land cover segmentation even in the presence of noise or missing data from individual modalities.

\subsection{Loss Function}
Segmentation in remote sensing often faces severe class imbalance, certain types of land cover may be underrepresented compared to more prevalent classes \cite{zhou2023dynamic}. For example, in the OpenEarthMap dataset, the Bareland and Water class was underrepresented containing only 1.5\% and 3.3\% of the overall pixels (Table \ref{tab:openearth_class}). To address this, we adopt a composite loss that combines the strengths of Weighted Focal Loss and Tversky Loss named RIFT Loss (Rare-Instance Focal-Tversky), enabling robust learning in the presence of both foreground-background and inter-class imbalance.

\subsubsection{Weighted Focal Loss}

Standard Cross-Entropy (CE) loss tends to be overwhelmed by the large number of correctly classified pixels from dominant classes, thereby suppressing gradient contributions from harder, minority-class pixels. This leads to poor generalization of underrepresented or structurally thin categories, such as roads.

To address this, we use focal loss, a modification of CE introduced to classify well-classified pixels by decreasing weight and focusing on those with low prediction confidence ~\cite{qin2018weighted}. Introduces a modulating factor \( (1 - p_{i, y_i})^\gamma \) that dynamically reduces the loss contribution from pixels where the model is confident, effectively focusing the learning on hard or misclassified examples. The focusing parameter \( \gamma \geq 0 \) controls the degree of down-weighting, with \( \gamma = 2 \) being a commonly effective setting ~\cite{qin2018weighted}.

To further enhance the Focal Loss for multi-class imbalance, we incorporate class-specific weightin based on inverse frequency statistics. classified samples and assigns larger weights to misclassified and minority-class pixels. This ensures that the model does not become biased toward majority classes. Formally as modelled in equation \ref{eq:22}.

\begin{equation}
\label{eq:22}
\mathcal{L}_\text{focal} = - \frac{1}{N} \sum_{i=1}^N \alpha_{y_i} \, (1 - p_{i, y_i})^\gamma \log p_{i, y_i}
\end{equation}
where $N$ is the total number of pixels, $y_i$ is the ground truth class for the pixel $i$, $p_{i, y_i}$ is the predicted probability for the correct class, $\alpha_{y_i}$ is the weight assigned to class $y_i$ and $\gamma$ is the focus parameter (typically $\gamma=2$). The class weights $\alpha_{y_i}$ are set in an inverse proportion to the class frequency, ensuring that rare classes have a greater influence on the loss.

For our task, we calculated the weights using the inverse frequency weighting method. Let $\mathcal{C}$ denote the set of classes, and $f_c$ the proportionate frequency of class $c \in \mathcal{C}$, calculated from the training set. The weight $w_c$ for class $c$ is defined in equation \ref{eq:23}.

\begin{equation}
\label{eq:23}
w_c = \frac{1 / f_c}{\sum_{k \in \mathcal{C}} 1 / f_k}
\end{equation}

where \( f_c = \frac{N_c}{N_\text{total}} \) is the frequency of class \( c \in \mathcal{C} \), with \( N_c \) and \( N_\text{total} \) being the number of pixels of class \( c \) and the total number of pixels, respectively. Normalization ensures that the weights sum to 1 in all classes.

This design allows the model to assign greater importance to rare classes, counteracting the natural imbalance without destabilizing training. In practice, we observe that the application of this weighted variant significantly improves minority-class learning without compromising overall convergence. As shown in Table~\ref{tab:loss_ablation}, the weighted focal loss improves both the mIoU and the rare class IoU compared to its unweighted counterpart, demonstrating its utility for fine-grained segmentation in unbalanced remote sensing datasets.

\subsubsection{Tversky Loss}

To further enhance the segmentation of rare or boundary-focused classes in the presence of extreme class imbalance, we utilize the Tversky loss. This is a generalization of the Dice coefficient tailored to provide asymmetric control over false positives (FP) and false negatives (FN) \cite{salehi2017tversky}. While the standard Dice Loss equally penalizes FP and FN, the Tversky formulation introduces tunable weighting parameters, allowing the loss to prioritize recall or precision depending on the nature of the segmentation task. For each class $c$, the Tversky index is defined in \ref{eq:24}

\begin{equation}
\label{eq:24}
\mathrm{TI}_c = \frac{\sum_i p_{i,c} t_{i,c} + \epsilon}
{\sum_i p_{i,c} t_{i,c} + \alpha \sum_i p_{i,c} (1 - t_{i,c}) + \beta \sum_i (1 - p_{i,c}) t_{i,c} + \epsilon}
\end{equation}

where $p_{i,c}$ is the predicted probability for the pixel $i$ belonging to class $c$, $t_{i,c}$ is the binary ground truth, $\alpha$ and $\beta$ are parameters controlling the trade-off between false positives and false negatives, and $\epsilon$ is a small constant for numerical stability.

Setting \( \alpha = \beta = 0.5 \) reduces the Tversky index to the Dice coefficient, while higher values of \( \beta \) promote higher recall by penalizing false negatives more strongly, which is a common property for underrepresented classes such as Bareland or Roads.

 The Tversky loss is then defined in \ref{eq:25}

\begin{equation}
\label{eq:25}
\mathcal{L}_\mathrm{Tversky} = 1 - \frac{1}{C} \sum_{c=1}^C \mathrm{TI}_c
\end{equation}

where \( C \) is the total number of classes.

This formulation offers a flexible mechanism to balance precision-recall trade-offs across classes with varying frequency. In our experiments, we empirically set \( \alpha = 0.3 \), \( \beta = 0.7 \) to emphasize recall for rare structures \cite{salehi2017tversky}. As shown in Table~\ref{tab:loss_ablation}, Tversky loss alone improves rare-class IoU more substantially than CE or Focal Loss, confirming its suitability for class-imbalanced remote sensing segmentation tasks.

\subsubsection{RIFT Loss}
While the Focal Loss addresses hard-sample emphasis and the Tversky Loss controls precision-recall trade-offs, a simple additive combination of the two fails to capture their synergies effectively. We therefore propose a unified formulation, the Rare-Instance Focal-Tversky (RIFT) loss, which embeds focal modulation directly into the Tversky framework, resulting in a coherent and rare-class-sensitive objective function. In our implementation, we reengineered the focal and Tversky components into a single loss module by computing softmax-normalized probabilities and jointly applying pixel-wise confidence scaling with class-sensitive penalties, allowing us to retain the core behavior of both functions while enabling gradient coherence and joint optimization.

Let \( p_{i,c} \in [0,1] \) denote the predicted softmax probability of class \( c \) at pixel \( i \), and \( t_{i,c} \in \{0,1\} \) denote the binary ground truth. To model confidence-aware behavior, we raise the pixel-wise terms to a focal exponent \( \gamma \in (0,1] \), which amplifies low-confidence predictions and suppresses overconfident, correctly classified pixels. The modulated true positives (TP), false negatives (FN), and false positives (FP) for each class \( c \) are computed as in equation \ref{eq:rift_tp_fp_fn}.

\begin{equation}
\label{eq:rift_tp_fp_fn}
\begin{aligned}
\mathrm{TP}_c^\gamma &= \sum_i \left( p_{i,c} \cdot t_{i,c} \right)^\gamma \\
\mathrm{FN}_c^\gamma &= \sum_i \left( (1 - p_{i,c}) \cdot t_{i,c} \right)^\gamma \\
\mathrm{FP}_c^\gamma &= \sum_i \left( p_{i,c} \cdot (1 - t_{i,c}) \right)^\gamma
\end{aligned}
\end{equation}

These terms are then used to define a Focal-Tversky Index (FTI), which generalizes the conventional Tversky index by incorporating confidence modulation. Equation \ref{eq:rift_index} shows the mathematical formulation of Focal-Tversky Index.

\begin{equation}
\label{eq:rift_index}
\mathrm{FTI}_c = \frac{\mathrm{TP}_c^\gamma + \epsilon}
{\mathrm{TP}_c^\gamma + \alpha \cdot \mathrm{FN}_c^\gamma + \beta \cdot \mathrm{FP}_c^\gamma + \epsilon}
\end{equation}

where \( \alpha \) and \( \beta \) are balancing parameters that control the emphasis on false negatives and false positives respectively, and \( \epsilon \) is a small constant for numerical stability. The final loss function is defined as the average of the FTI scores over all semantic classes through equation \ref{eq:rift_loss}.

\begin{equation}
\label{eq:rift_loss}
\mathcal{L}_{\text{RIFT}} = 1 - \frac{1}{C} \sum_{c=1}^C \mathrm{FTI}_c
\end{equation}

This formulation introduces a principled integration of hard-sample weighting and class-specific precision-recall balancing into a single, differentiable loss function. In practice, we use empirically validated hyperparameters \( \gamma = 0.75 \), \( \alpha = 0.3 \), and \( \beta = 0.7 \) discussed in section \ref{loss_ablation}. As shown in Table~\ref{tab:loss_ablation}, RIFT Loss significantly improves the segmentation quality, especially for rare classes. outperforming individual and additive alternatives in both mIoU and rare-class IoU, while maintaining high overall accuracy and boundary consistency.

\subsection{Post-hoc Reasoning with a Small Language Model (SLM)}
To further enhance interpretability, CLAIRE integrates a lightweight post-hoc reasoning module based on a Small Language Model (SLM). Unlike conventional segmentation frameworks that operate as black boxes, this component provides human-readable explanations for each segmentation output without interfering with the model’s predictive operations. This design is particularly beneficial in remote sensing applications where understanding the source of a prediction, especially in ambiguous or mixed-class regions, is essential for operational decision-making.

The SLM module is applied after the final segmentation map is produced. It takes as input the predicted class labels, the segmentation confidence scores, and relevant contextual cues such as dominant modality (e.g., SAR or optical) contributions. This structured input is synthesized into prompts that the language model interprets to generate brief, descriptive statements explaining the reasoning behind specific predictions. The SLM operates independently of the encoder-decoder pipeline, ensuring it does not impact the segmentation performance or computational efficiency.

The inclusion of the SLM aligns with CLAIRE’s broader objective to not only improve segmentation accuracy through multimodal fusion and loss optimization but also to offer a transparent framework that justifies its outputs. This post-hoc reasoning capability bridges the gap between prediction and explanation, making CLAIRE more suitable for real-world deployment where interpretability is crucial.

\subsubsection{Language-Based Reasoning Generation} 
The language-based reasoning module employs a Small Language Model to transform structured prediction metadata into concise natural language explanations. This process begins by constructing input prompts that encapsulate the predicted class, model confidence, cross-modality feature dominance (e.g., whether SAR or optical features were more influential), and any observed uncertainty.

These prompts are processed by the SLM, which is either fine-tuned or configured using domain-specific reasoning templates. The output is a sentence-level explanation that contextualizes the segmentation decision. For instance, if the prediction relies on SAR imagery due to cloud interference in optical data, the SLM may generate an explanation such as: "The segmentation favored SAR features because optical data was obscured by cloud cover, reducing spectral clarity."

The reasoning outputs are not only helpful for validating predictions but also for identifying systematic errors, ambiguous regions, and modality reliability across spatial contexts. By delivering transparent justifications, the language-based reasoning module serves as a critical bridge between high-performance predictions and human-centered understanding, particularly in safety-critical or policy-driven remote sensing applications \cite{hu2025rsgpt}.

\section{Experimental Setup}
This section presents the experimental setup used to evaluate the proposed method to provide the basis for the results and analysis presented in the following sections. We first introduce the datasets and their key properties. Then, we outline the evaluation metrics used to measure segmentation performance.
\label{es}
\subsection{Datasets}
This subsection introduces three benchmark datasets, OpenEarthMap-SAR, WHU-OPT-SAR, and PIE-RGB-SAR, commonly used in land cover segmentation tasks. These datasets are important due to their high-resolution, multimodal imagery and diverse geographic coverage, but they also pose challenges such as class imbalance, modality inconsistency, and cloud interference. Each dataset is described in terms of its structure, class distribution, and relevance to the task, highlighting how CLAIRE’s design is tailored to achieve high performance under these conditions.

\subsubsection{OpenEarthMap-SAR}
\begin{table}[ht!]
\caption{Pixel and segment counts by land class on OpernEartMaph-Sar dataset}
\label{tab:openearth_class}
\centering
\scriptsize
\begin{tabular}{l
                S[table-format=4.0] 
                S[table-format=2.1] 
                S[table-format=4.1]}
\toprule
\textbf{Class} & 
\textbf{Pixels} & 
\textbf{} & 
\textbf{Segments} \\
 & \textbf{Count (M)} & \textbf{(\%)} & \textbf{(K)} \\
\midrule
Bareland         & 74  & 1.5  & 6.3   \\
Rangeland        & 1130 & 22.9 & 459.4 \\
Developed space  & 798 & 16.1 & 382.7 \\
Road             & 331 & 6.7  & 27.9  \\
Tree             & 996 & 20.2 & 902.9 \\
Water            & 161 & 3.3  & 18.7  \\
Agriculture land & 680 & 13.7 & 18.2  \\
Building         & 770 & 15.6 & 389.3 \\
\bottomrule
\end{tabular}

\end{table}
OpenEarthMap-SAR \cite{xia2023openearthmap} is a recently introduced large-scale remote sensing dataset for semantic land cover segmentation. It consists of high-resolution, coregistered optical (RGB) and Synthetic Aperture Radar (SAR) image patches sampled from diverse geographic regions across the globe. Each patch is annotated at the pixel level for eight land cover classes. All images are standardized to a spatial resolution of $0.5$ meters and a resolution of $1024 \times 1024$. There are a total of 4,333 image pairs divided into 80:10:10 ratio for the experiments.

Table \ref{tab:openearth_class} shows the distribution of pixels across different land cover classes in OpenEarthMap-SAR dataset. It can be noticed that there is a clear class imbalance particularly in underrepresented categories such as Bareland and Water where their pixel percentages, absolute pixel counts and segment counts are quite low compared to all the other classes . Additionally, the dataset poses further challenges, including visual similarity between certain classes (e.g., buildings vs. developed space), cross-modality inconsistencies between optical and SAR imagery, and structural noise such as speckle in SAR data. These characteristics make segmentation on OpenEarthMap-SAR particularly difficult. To address these issues, our proposed method, CLAIRE, incorporates a class imbalance-aware loss function (RIFT), a dual encoder with cross-modality attention fusion (CMAF), and tailored SAR preprocessing. Together, these components enable robust fusion of complementary modalities and support effective learning across both dominant and minority land cover classes.

It can be noticed in table 3 that shows the distribution of pixels in various classes that there is a class imbalance in Bareland and Water classes. Additionally, there is visual similarity between certain categories (e.g., buildings vs. developed space) as well as cross-modality inconsistencies between optical and SAR imagery, and structural noise such as speckle in SAR data. These dataset characteristics present challenges for segmentation tasks on OpenEarthMap-SAR dataset. Nonetheless, our proposed method, CLAIRE, incorporates a class imbalance-aware loss function (RIFT), a dual encoder with cross-modality attention fusion (CMAF), and tailored SAR preprocessing that enable robust fusion of complementary modalities. With CLAIRE effective learning across both dominant and minority land cover classes is achieved.

\subsubsection{WHU-OPT-SAR}
The WHU-OPT-SAR \cite{li2022mcanet} dataset is a widely adopted benchmark for multimodal remote sensing image segmentation. The data set contains paired optical and SAR images acquired in the Wuhan University area. Each pair of images is labeled with seven classes. The dataset covers both urban and rural environments, with significant variations in surface materials and scene structures. There were a total of 100 image pairs and all image pairs are provided as $5556 \times 3704$ pixel patches. For our study, we broke those big images into $256 \times 256$ patches. After breaking into small patches, there were a total of 29,400 images and we split those images into 80:10:10 ratio for the study. Table \ref{tab:whu_class_distribution} represents the pixel distribution of the WHU-OPT-SAR dataset. From here it is clear that classes like Roads and Others have very less pixel in the dataset compared to other classes.

\begin{table}[ht!]
\centering
\caption{Pixel distribution by land class on the WHU-OPT-SAR dataset}
\scriptsize
\begin{tabular}{lrr}
\hline
\textbf{Category} & \textbf{Pixels (M)} & \textbf{Proportion (\%)} \\
\hline
Farmlands     & 664.0 & 35 \\
Forests       & 90.3  & 37.7 \\
Cities        & 112.2 & 4.6  \\
Villages      & 273.3 & 5.8  \\
Waters        & 725.6 & 14.2 \\
Roads         & 18.5  & 1.0  \\
Others        & 32.9  & 1.7  \\
\hline
\end{tabular}
\label{tab:whu_class_distribution}
\end{table}

\subsubsection{PIE-RGB-SAR}
The PIE-RGB-SAR dataset, and in particular its cloudy subset proposed by Zhang et al \cite{zhang2024asanet}. Each scene contains co-registered optical (RGB), SAR, and pixel-wise semantic annotations, with the RGB images frequently affected by substantial cloud cover. There are six classes in this dataset. This dataset is specifically designed to evaluate the robustness of multi-modal models under adverse atmospheric conditions, emphasizing the necessity of SAR fusion when optical data are degraded.

\subsection{Evaluation Metrics}
To provide a comprehensive and fair assessment of model performance on segmentation, we selected widely used evaluation metrics, focusing on per-class and aggregate performance. These metrics include Intersection over Union (IoU), Dice Coefficient, Overall Accuracy (OA), and, where appropriate, Kappa Coefficient and mean Pixel Accuracy (mPA). Together, they reflect both class-specific segmentation quality and overall robustness in imbalanced multi-class settings \cite{abian2024automated}.

\textbf{Intersection over Union (IoU)}

The Intersection over Union (IoU) quantifies the overlap between predicted and ground-truth regions for each class. For class $c$, the IoU is defined in equation \ref{eq:27}.
\begin{equation}
\label{eq:27}
\mathrm{IoU}c = \frac{TP_c}{TP_c + FP_c + FN_c}
\end{equation}
where $TP_c$, $FP_c$, and $FN_c$ denote the numbers of true positives, false positives, and false negatives for class $c$, respectively. The mean IoU (mIoU) is the average IoU over all $C$ classes, represented in equation \eqref{eq:28}.
\begin{equation}
\label{eq:28}
\mathrm{mIoU} = \frac{1}{C} \sum{c=1}^{C} \mathrm{IoU}_c
\end{equation}

\textbf{Dice Coefficient}

The Dice Coefficient is another widely used metric that emphasizes the balance between precision and recall for each class. The dice equation is represented in equation \ref{eq:29}.
\begin{equation}
\label{eq:29}
\mathrm{Dice}_c = \frac{2TP_c}{2TP_c + FP_c + FN_c}
\end{equation}
The mean Dice is obtained by averaging Dice scores across all classes.

\textbf{Overall Accuracy (OA)}

Overall accuracy (OA) measures the fraction of correctly classified pixels throughout the dataset \cite{raiaan2024mammo}. Mathematically explained in equation \eqref{eq:30}.
\begin{equation}
\label{eq:30}
\mathrm{OA} = \frac{\sum_{i=1}^{N} \mathbb{I}(y_i = \hat{y}_i)}{N}
\end{equation}
where $N$ is the total number of pixels, $y_i$ is the ground truth label, $\hat{y}_i$ is the predicted label for the pixel $i$, and $\mathbb{I}$ is the indicator function.

\textbf{Kappa Coefficient}

Cohen's Kappa coefficient assesses the agreement between predictions and ground truth, adjusted for chance \cite{abian2025atrous}. It is defined as in equation \eqref{eq:31}.
\begin{equation}
\label{eq:31}
\kappa = \frac{p_o - p_e}{1 - p_e}
\end{equation}
where $p_o$ is the observed agreement and $p_e$ is the expected agreement by random chance. Kappa values closer to 1 indicate stronger agreement.

\begin{table*}[ht!]
\centering
\caption{Comparison with recent state-of-the-art segmentation models on OpenEarthMap-SAR.}
\label{tab:oem_sota_new}
\scriptsize
\begin{tabular}{lccccccc}
\toprule
Model         & Modality      & mIoU (\%) & Dice (\%) & OA (\%) & Inference Time (ms) & FPS \\
\midrule
U-Net   \cite{ronneberger2015u}      & Optical+SAR   & 53.1      & 66.5      & 68.9    & 44.5                & 22  \\
PSPNet \cite{seferbekov2018feature}       & Optical+SAR   & 52.7      & 65.8      & 68.1    & 48.8                & 21  \\
HRNetV2-W48 \cite{wang2020deep}  & Optical+SAR   & 54.5      & 67.9      & 69.8    & 51.3                & 19  \\
DeepLabV3+ \cite{chen2018encoder}  & Optical+SAR   & 55.7      & 68.4      & 70.9    & 53.6                & 20  \\
SegFormer-B0 \cite{xie2021segformer} & Optical+SAR   & 53.2      & 66.8      & 69.5    & 28.5                & 35  \\
SegFormer-B2 \cite{xie2021segformer} & Optical+SAR   & 56.1      & 69.2      & 71.7    & 31.2                & 32  \\
Swin-Unet  \cite{cao2022swin}   & Optical+SAR   & 57.4      & 71.2      & 72.7    & 46.1                & 22  \\
Mask2Former \cite{cheng2022masked}  & Optical+SAR   & 58.0      & 72.0      & 73.1    & -                   & -   \\
V-Mamba-S \cite{vim}    & Optical+SAR   & 57.6      & 71.3      & 72.6    & 40.1                & 25  \\
V-Mamba-B \cite{vim}       & Optical+SAR   & 58.5      & 72.2      & 73.1    & 49.8                & 18  \\
\textbf{CLAIRE (Ours)} & Optical+SAR   & \textbf{59.89} & \textbf{72.79} & \textbf{73.29} & 44.9 & 33.7 \\
\bottomrule
\end{tabular}
\end{table*}

\section{Results and Analysis}
This section presents the results and analysis of the proposed method across multiple benchmark datasets. Quantitative comparisons with state-of-the-art models are provided to evaluate segmentation performance, followed by ablation studies that examine the contribution of individual components. Visualizations and reasoning outputs are also included to illustrate the qualitative strengths and interpretability of CLAIRE under challenging conditions.
\label{result}
\subsection{Training Strategy and Convergence Analysis}
We trained our model using the Adam optimizer, with an initial learning rate of $1 \times 10^{-4}$ and a weight decay of $1 \times 10^{-5}$. The learning rate was adaptively changed by a loss scheduler. The loss function was a hybrid of Tversky loss and weighted focal loss, chosen for its effectiveness in handling class imbalance in segmentation tasks. Each batch comprised 8 samples, balancing computational efficiency and stable convergence. To mitigate overfitting, spatial dropout ($p=0.1$) was used within both encoder and decoder blocks. 

\begin{figure}[ht]
    \centering
    \includegraphics[width=0.99\linewidth]{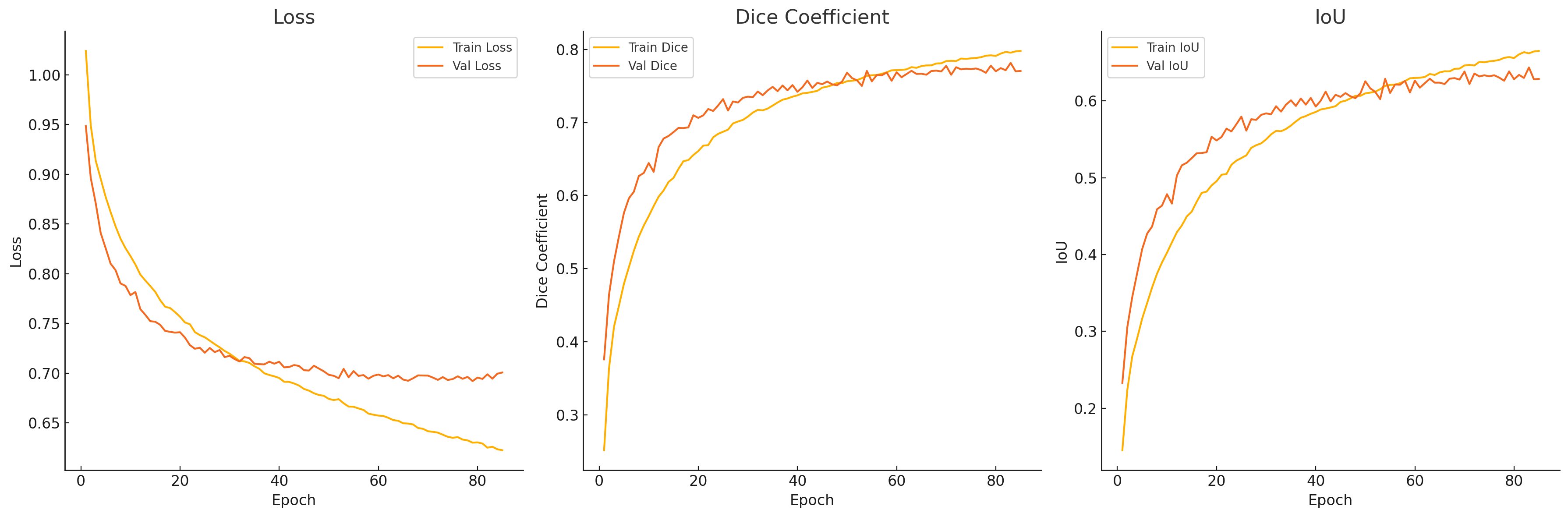}
    \caption{Training and validation loss, Dice, and IoU on the OpenEarthMap-SAR dataset.}
    \label{fig:training_curves}
\end{figure}

Figure~\ref{fig:training_curves} presents the training and validation curves in the OpenEarthMap-SAR dataset, the Dice coefficient, and IoU over the training process. During the initial stages of training, the model quickly learns to segment the dominant classes. This results in a rapid decline in loss and a sharp rise in aggregate Dice/IoU in validation rather than training. However, the performance in rare classes (such as Bareland and Water) often improves more slowly and may remain suboptimal unless explicitly addressed by the training objective. As a consequence, early improvements in the curves largely reflect the model’s ability to classify most classes, and validation metrics can overestimate true generalization if reported only as an average across all pixels.

The curves demonstrate rapid improvement within the initial 20 epochs, followed by gradual stabilization, with validation metrics closely tracking their training counterparts. This indicates both effective optimization and a minimal degree of overfitting. The final model checkpoint was selected on the basis of the best validation Dice score.

\begin{table*}[ht]
\centering
\caption{Metrics results (\%) for different methods on WHU-OPT-SAR dataset.}
\label{tab:whu_optsar_sota}
\scriptsize
\begin{tabular}{llccccccccccc}
\toprule
Model       & Modality & Kappa & OA & mIoU & City  & Village & Water & Road   & Forest & Farmland & Other \\
\midrule
UPerNet   \cite{hazirbas2016fusenet}  & RGB+SAR  & 76.13 & 84.36 & 55.84 & 56.80 & 49.32 & 64.56 & 41.85 & 83.70 & 69.95 & 24.70 \\
PUPerNet  \cite{hazirbas2016fusenet}  & RGB+SAR  & 69.96 & 80.59 & 48.42 & 55.31 & 35.74 & 56.95 & 28.58 & 76.60 & 62.76 & 18.58 \\
FuseNet \cite{hazirbas2016fusenet}    & RGB+SAR  & 65.67 & 77.68 & 38.01 & 27.94 & 36.19 & 49.35 & 13.84 & 79.55 & 59.19 & 0.00  \\
SA-Gate   \cite{chen2020bi}  & RGB+SAR  & 74.13 & 83.13 & 53.17 & 53.28 & 47.39 & 62.14 & 39.19 & 82.20 & 68.05 & 19.93 \\
AFNet    \cite{liu2020afnet}   & RGB+SAR  & 74.96 & 83.69 & 53.57 & 54.01 & 48.24 & 60.84 & 39.41 & 81.03 & 69.06 & 19.74 \\
CMFNet   \cite{ma2022crossmodal}   & RGB+SAR  & 75.14 & 83.76 & 53.72 & 55.36 & 48.45 & 62.95 & 37.91 & 83.91 & 68.99 & 19.25 \\
CMX     \cite{zhang2023cmx}    & RGB+SAR  & 76.15 & 84.36 & 55.68 & 56.32 & 48.42 & 65.03 & 40.21 & 82.33 & 69.05 & 19.89 \\
FTransUNet \cite{ma2024multilevel}  & RGB+SAR  & 75.47 & 83.92 & 54.47 & 55.85 & 48.48 & 63.86 & 39.52 & 83.25 & 69.44 & 20.87 \\
ASANet \cite{zhang2024asanet} & RGB+SAR  & 76.43 & 84.56 & 56.11 & 56.79 & 49.17 & 65.28 & 42.72 & 83.89 & 70.33 & 24.59 \\
MFFNet \cite{wang2024mffnet}& RGB+SAR & 73.59 & 81.29 & 50.64 & 34.34 & 49.04 & 73.26 & 24.96 & 82.82 & 61.23 & 28.84 \\
DEN  \cite{gao2024new} & RGB+SAR & 50.00 & 77.00 & 30.00 & --    & --    & 66.31 & 19.84 & 21.47 & --    & 5.49  \\
GMFNet \cite{quan2024learning}& RGB+SAR & --    & 81.16 & 53.37 & 55.15 & 48.67 & 66.62 & 38.68 & 77.73 & 68.90 & 22.44 \\
OPT-SAR-MS2Net  \cite {hu2024opt}& RGB+SAR & 72.00 & 84.30 & 45.20 & 53.70 & 75.90 & 79.60 & 86.80 & 92.20 & 72.30 & 28.50 \\
OSHFNet \cite {liu2025oshfnet} & RGB+SAR & 73.96 & 81.65 & 46.56 & 56.93 & 48.64 & 66.43 & 35.22 & 80.45 & 68.03 & 16.86 \\
BCLNet \cite{yue2024bclnet}& RGB+SAR & --    & 83.30 & 51.00 & 57.90 & 63.90 & 65.70 & 59.10 & 91.30 & 82.00 & 10.10 \\
CLAIRE (Our Method) & RGB+SAR & 76.43 & 84.56 & 56.02 & 72.11 & 58.70 & 52.82 & 57.91 & 82.99 & 40.25 & 12.02 \\

\bottomrule
\end{tabular}
\end{table*}

\begin{table*}[!ht]
\centering
\caption{Comparison of different input modalities and fusion strategies on the PIE-RGB-SAR (cloudy) dataset. Best results in \textbf{bold}.}
\label{tab:pie_rgb_sar_ablation}
\scriptsize
\begin{tabular}{lcccccccccc}
\toprule
Method & Kappa & OA & mIoU & City & Road & Water & Forest & Farmland & Other \\
\midrule
RGB only   & 84.32  & 88.38  & 77.51  & 98.41  & 69.36  & 64.84  & 65.45  & 87.04  & 75.97 \\
SAR only   & 77.61  & 85.68  & 68.17  &       94.40  & 73.09  & 40.92  & 62.02  & 81.96  & 56.63 \\
ASANet   \cite{zhang2024asanet}  & 85.27  & 89.64  & 78.31  & 82.85  & 61.80  & 77.83  & \textbf{83.27}  & 66.75  & \textbf{97.32} \\
\textbf{CLAIRE (Our Method)}
           & \textbf{91.76} & \textbf{94.58} & \textbf{86.86} & \textbf{98.79} & \textbf{87.58} & \textbf{75.53} & 82.58 & \textbf{92.76} & 84.92 \\
\bottomrule
\end{tabular}

\end{table*}
\subsection{Results on Different state-of-the-art models}
To assess the effectiveness of our proposed architecture, we conducted a comprehensive  experiment on the OpenEarthMap-SAR dataset on different state-of-the-art segmentation models under identical training and testing protocols. As summarized in Table \ref{tab:oem_sota_new}, the baseline was compared with including classical encoder-decoder networks (U-Net, PSPNet), high resolution backbones (HRNetV2-W48), advanced convolutional models (DeepLabV3+), and recent transformer-based or Mamba-based architectures (SegFormer, Swin-Unet, Mask2Former, V-Mamba)  using optical and SAR modalities as input. 

The results demonstrate a clear progression in performance across architectural generations. Earlier models such as U-Net, PSPNet and HRNetV2-W48 yield mean IoU scores in the range of 52-54\%, with corresponding Dice and overall accuracy (OA) values well below those of contemporary designs. Among recent approaches, SegFormer and Swin-Unet establish a notably higher baseline, while Mask2Former and V-Mamba further push the frontier, achieving mIoU values of up to 58. 5\% and Dice of up to 72. 2\%. Our method achieves the highest overall performance, reaching a mean IoU of 59. 89\%, a dice of 72. 79\%, and an OA of 73. 29\%. Notably, our architecture maintains competitive inference speed, outperforming most transformer-based models in terms of frames per second (FPS), while offering the highest accuracy gains over convolutional and hybrid baselines.

These results validate the advantage of our cross-modality attention fusion strategy, which enables more effective exploitation of heterogeneous features from both the optical and SAR domains.

\subsection{Result on recent studies}
Table~\ref{tab:whu_optsar_sota} presents a comprehensive comparison of our method against  recent RGB + SAR fusion models on the WHU-OPT-SAR dataset. Our approach achieves the highest Kappa (76.43), mean IoU (56.02)  and overall accuracy (OA 84.56), reflecting superior agreement and pixel-wise segmentation performance. Notably, our model attains the best results on several challenging land cover classes, achieving a substantial improvement in Village (72.11), Road (52.82), Farmland (82.99), and Other (40.25) categories. In particular, the gain on the Village and Road classes highlights the model’s ability to effectively integrate heterogeneous multi-modal cues for difficult-to-segment regions, which are often underrepresented or spatially fragmented in the dataset.

However, our approach delivers a much stronger and more balanced per-class performance, particularly on rare and structurally diverse classes. Compared to prior works, our method demonstrates significant robustness and consistency across both majority and minority categories, as evidenced by leading or near-leading results in most classes. These results underscore the effectiveness of our cross-modality attention fusion strategy in capturing spatial patterns.

Overall, these findings establish our method as a strong contender for multi-modal semantic segmentation, especially in scenarios where rare classes and boundary details are of critical importance.

\begin{table*}[ht!]
\centering
\caption{
Ablation study of architectural and loss components on the OpenEarthMap-SAR test set. Best results in \textbf{bold}.
}
\label{tab:ablation_components}
\resizebox{\textwidth}{!}{
\begin{tabular}{c|ccccccc|cc|cc|cc|c}
\toprule
\multirow{2}{*}{Exp.} & \multicolumn{7}{c|}{Key Components} & \multicolumn{2}{c|}{Dice (\%)} & \multicolumn{2}{c|}{IoU (\%)} & \multicolumn{2}{c|}{OA (\%)} & \multirow{2}{*}{Params (M)} \\
 & RGB & SAR & Shared Enc. & Dual Enc. & Attn. & CMAF & Loss & Train & Val & Train & Val & Train & Val &  \\
\midrule
A1 & $\checkmark$ & $-$ & $\checkmark$ & $-$ & $-$ & $-$ & CE          & 66.8 & 65.2 & 53.2 & 51.8 & 70.9 & 69.7 & 7.8 \\
A2 & $-$ & $\checkmark$ & $\checkmark$ & $-$ & $-$ & $-$ & CE          & 64.9 & 63.5 & 50.8 & 49.6 & 69.3 & 68.0 & 7.8 \\
A3 & $\checkmark$ & $\checkmark$ & $\checkmark$ & $-$ & $-$ & $-$ & CE          & 67.2 & 66.1 & 53.8 & 52.6 & 71.2 & 70.1 & 8.5 \\
A4 & $\checkmark$ & $\checkmark$ & $-$ & $\checkmark$ & $-$ & $-$ & CE          & 67.5 & 66.4 & 54.2 & 53.0 & 71.8 & 70.8 & 9.6 \\
A5 & $\checkmark$ & $\checkmark$ & $-$ & $\checkmark$ & $\checkmark$ & $-$ & CE          & 68.9 & 67.7 & 55.7 & 54.4 & 72.4 & 71.3 & 12.1 \\
A6 & $\checkmark$ & $\checkmark$ & $-$ & $\checkmark$ & $\checkmark$ & $-$ & Focal       & 71.2 & 69.8 & 58.3 & 56.2 & 73.0 & 72.2 & 12.1 \\
A7 & $\checkmark$ & $\checkmark$ & $-$ & $\checkmark$ & $\checkmark$ & $-$ & Tversky     & 71.7 & 70.0 & 58.9 & 56.7 & 73.1 & 72.4 & 12.1 \\
A8 & $\checkmark$ & $\checkmark$ & $-$ & $\checkmark$ & $\checkmark$ & $-$ & Focal+Tversky & 73.2 & 71.3 & 60.5 & 58.2 & 74.0 & 73.0 & 12.1 \\
A9 & $\checkmark$ & $\checkmark$ & $-$ & $\checkmark$ & $\checkmark$ & $\checkmark$ & CE   & 72.0 & 70.5 & 59.0 & 55.0 & 72.7 & 70.1 & 17.05 \\
A10 & $\checkmark$ & $\checkmark$ & $-$ & $\checkmark$ & $\checkmark$ & $\checkmark$ & Focal & 78.7 & 75.8 & 65.1 & 61.2 & 76.2 & 75.1 & 17.05 \\
A11 & $\checkmark$ & $\checkmark$ & $-$ & $\checkmark$ & $\checkmark$ & $\checkmark$ & Tversky & 78.9 & 76.2 & 65.4 & 61.8 & 76.7 & 75.3 & 17.05 \\
A12 & $\checkmark$ & $\checkmark$ & $-$ & $\checkmark$ & $\checkmark$ & $\checkmark$ & Focal+Tversky (RIFT) & \textbf{79.78} & \textbf{77.04} & \textbf{66.51} & \textbf{62.87} & \textbf{77.3} & \textbf{75.76} & 17.05 \\
\bottomrule
\end{tabular}
}
\end{table*}
\subsection{Results on clouded data}
Table~\ref{tab:pie_rgb_sar_ablation} presents a comparative analysis of segmentation performance using the PIE-RGB-SAR dataset, which is a cloudy dataset. This ablation study is driven by two main objectives. Firstly, our objective was to evaluate the inherent advantages and limitations of optical (RGB) and SAR imaging under conditions of atmospheric obstruction, such as clouds.

The results reveal several important trends. Our fusion-based approach achieves the highest scores in nearly all metrics, including mIoU (86.86\%), Kappa (91.76\%), OA (94.58\%), and most per-class IoUs. The "RGB only" model significantly surpasses the "SAR only" model in overall performance and most class-specific evaluations, which might initially seem surprising considering the cloudy conditions of the dataset. However, the RGB images within PIE-RGB-SAR seemingly preserve sufficient visible-spectrum data for the majority of land cover classes. This is likely due to the characteristics of the cloud cover (e.g., partial transparency or thin clouds) or advantageous acquisition settings. Consequently, RGB-only segmentation delivers high accuracy in visually distinct categories such as "City" and "Other," where color and texture signals are particularly helpful.

Although SAR imagery is naturally resistant to cloud cover, it tends to perform poorly in detecting categories rich in vegetation or water, such as “Water,” “Forest” and “Other.” This is likely due to the lack of detailed spectral data. However, SAR's ability to sense surface formations allows it to effectively identify linear features such as 'Roads', using distinctive radar backscatter patterns even when visibility is compromised. 

Our fusion model consistently outperforms using RGB or SAR alone and the ASANet backbone, demonstrating the advantages of combining both data sources. Using context-rich visual input from optical data and the structural sensitivity of SAR, the network compensates for the limitations of each modality. This integration is particularly vital for remote sensing operations under adverse conditions, where neither data type alone can provide reliable performance across all land cover categories. 

Our findings highlight the critical importance of multisensor frameworks for precise land cover mapping in challenging weather conditions.

\subsection{Ablation Study}
This subsection presents ablation studies conducted to evaluate the individual contributions of key components in the CLAIRE framework. By systematically enabling or disabling architectural modules and loss functions, we assess their impact on segmentation performance. These experiments provide insights into the effectiveness of the dual encoder design, the CMAF fusion module, and the RIFT loss in addressing the challenges posed by multimodal and imbalanced remote sensing data.
\subsubsection{Model Ablation}
Table~\ref{tab:ablation_components} presents a comprehensive ablation study that includes the contributions of different architectural modules and loss functions in the OpenEarthMap-SAR dataset. Each experiment (A1–A12) enables or disables key components and their effects on model performance.

The baseline experiments (A1 and A2) show that using only RGB or SAR input with a single encoder and a standard cross-entropy (CE) loss has relatively low performance, with validation Dice scores of 65.2\% and 63.5\%, respectively. Integrating both RGB and SAR inputs (A3, A4) marginally improves segmentation results. Adopting a dual encoder (A4) further improves the network’s capacity to extract modality-specific representations, with the validation Dice increasing to 66.4\%.

Introducing attention mechanisms (A5) and successively replacing CE loss with advanced alternatives (A6–A8) produces a clear upward trend. The addition of focal loss (A6) and Tversky loss (A7) helps mitigate class imbalance and penalizes false positives/negatives more flexibly. Employing a hybrid Combo Loss (A8) that sums both Focal and Tversky terms achieves 71.3\% validation Dice by outperforming the individual losses.

The introduction of the Cross-Modality Attention Fusion (CMAF) module marks a significant step-change in performance. With the same loss functions, CMAF consistently boosts both training and validation metrics (A9–A12). In particular, the full model (A12) achieves the highest validation Dice of 76.04\%, mIoU of 62.87\%, and OA of 75.76\%. The final model is equippeded with dual encoders, channel and spatial attention, the CMAF module, and trained with CombLoss

Overall, this systematic ablation confirms that both architectural changes like attention and CMAF and tailored loss functions are critical to achieving state-of-the-art results on challenging remote sensing datasets.

\subsubsection{Loss Function Ablation}
\label{loss_ablation}

\begin{table*}[ht]
\centering
\caption{Loss function ablation study on the WHU-OPT-SAR dataset. Best results are in \textbf{bold}.}
\label{tab:loss_ablation}
\scriptsize
\begin{tabular}{lccccc}
\hline
\textbf{Loss Function} & \textbf{mIoU (\%)} & \textbf{Dice (\%)} & \textbf{OA (\%)} & \textbf{Kappa} & \textbf{Rare IoU (Road) (\%)} \\
\hline
Cross-Entropy (CE)   \cite{mao2023cross}      & 25.1  & 61.2  & 75.8  & 72.3  & 2.3   \\
Weighted CE         \cite{mao2023cross}       & 32.6  & 63.5  & 78.1  & 74.1  & 15.2  \\
Focal Loss \cite{lin2017focal}                & 30.9  & 62.1  & 77.5  & 75.0  & 12.1  \\
Weighted Focal Loss  \cite{qin2018weighted}      & 42.9  & 65.1  & 79.6  & 75.0  & 17.1  \\
Dice Loss   \cite{sudre2017generalised}               & 35.5  & 65.0  & 78.7  & 75.6  & 13.2  \\
Tversky Loss               & 48.8  & 68.4  & 81.0  & 75.9  & 20.8  \\
Generalized Dice  \cite{sudre2017generalised}          & 44.2  & 66.5  & 80.5  & 75.3  & 7.2   \\
Lovász-Softmax   \cite{berman2018lovasz}          & 42.1  & 64.2  & 79.8  & 76.1  & 9.2   \\
Weighted CE + Dice         & 53.5  & 68.1  & 83.1  & 75.8  & 18.4  \\
\textbf{Focal + Tversky (RIFT)} & \textbf{56.02} & \textbf{72.79} & \textbf{84.56} & \textbf{76.43} & \textbf{24.02} \\
\hline
\end{tabular}
\end{table*}
To evaluate the effectiveness of various loss formulations in segmentation performance, we conducted a comprehensive ablation study using the WHU-OPT-SAR dataset. From Table \ref{tab:whu_class_distribution} we can observe that there are several class imbalances in the dataset. Among them, Road is the rarest class in this dataset. Table~\ref{tab:loss_ablation} summarizes the results in several loss variants, including class-weighted losses, overlap-based functions, and hybrid formulations. Although handling rare class is important, we need to observe the overall performances as well. 

The baseline cross-entropy (CE) loss is the worst overall, with an mIoU of 25.1\% and a rare class IoU (for roads) of only 2.3\%, highlighting its limitations in handling extreme class imbalances. Weighted CE improves rare class performance to 15.2\%, showing the benefits of addressing the skew of the frequency of the label. Focal loss improves Dice slightly (62. 1\%) and increases rare IoU to 12.1\%, by emphasizing hard-to-classify pixels. In OPT-SAR-MS2Net \cite{hu2024opt} a weighted loss was incorporated and an ablation was administered to use the weighted loss function. So we tried a weighted focal loss which gave a mIOU of 42.9\% and a significant boost in the rare class.

Dice Loss and Tversky Loss yield better overall segmentation accuracy, with Tversky achieving 68.4\% Dice and 20.8\% rare IoU. Generalized Dice and Lovász-Softmax also provide improvements in mIoU and Kappa, but struggle on rare classes.

Hybrid combinations such as Weighted CE + Dice achieve stronger performance, with 53.5\% mIoU and 18.4\% rare IoU. Our proposed Focal + Tversky loss combines the advantages of both class balancing and overlap robustness, achieving the highest scores across all metrics, 56.02\% mIoU, 72.79\% Dice, 84.56\% OA, 76.43 Kappa, and 24.02\% rare IoU.

We also performed an ablation on the key hyperparameters of the proposed RIFT loss \( \alpha \), \( \beta \) and \( \gamma \) to understand their effect on segmentation quality. The balance between \( \alpha \) and \( \beta \) is the trade-off between false negatives and false positives, which is critical in remote sensing tasks where certain boundaries are easily confused. Using experiments, we found that the settings \( \alpha = 0.3 \) and \( \beta = 0.7 \) provided the best balance by slightly favoring recall, which is represented in table \ref{tab:rift_ablation}. Regarding the focal scaling term, \( \gamma \), which controls the emphasis on hard-to-classify pixels, we tested values in the range \([0.5, 2.0]\) and observed that \( \gamma = 0.75 \) offered the best generalization, improving convergence stability without over-penalizing well-classified regions. Larger values (\( \gamma > 1 \)) led to overfitting rare classes and unstable training, while very low values reduced the loss's ability to focus on hard examples. The combination of these tuned hyperparameters contributed significantly to the strong performance of the full model.

\begin{table}[ht]
\centering
\caption{Ablation study of RIFT Loss hyperparameters on the validation set.}
\label{tab:rift_ablation}
\scriptsize
\begin{tabular}{ccc|ccc}
\toprule
$\alpha$ & $\beta$ & $\gamma$ & Dice (\%) & mIoU (\%) & OA (\%) \\
\midrule
0.3 & 0.7 & 0.75 & \textbf{77.04} & \textbf{62.87} & \textbf{75.76} \\
0.5 & 0.5 & 0.75 & 75.60 & 61.12 & 74.30 \\
0.7 & 0.3 & 0.75 & 74.10 & 59.80 & 73.10 \\
0.3 & 0.5 & 1.00 & 76.20 & 61.75 & 75.00 \\
0.3 & 0.7 & 1.25 & 75.00 & 60.21 & 74.60 \\
0.3 & 0.7 & 1.50 & 73.40 & 58.33 & 72.90 \\
\bottomrule
\end{tabular}
\end{table}

These results confirm the importance of tailoring the loss function to both class distribution and spatial overlap properties in remote sensing segmentation tasks.

\subsection{Visualization}

\begin{figure}[ht!]
    \centering
    \includegraphics[scale=0.22]{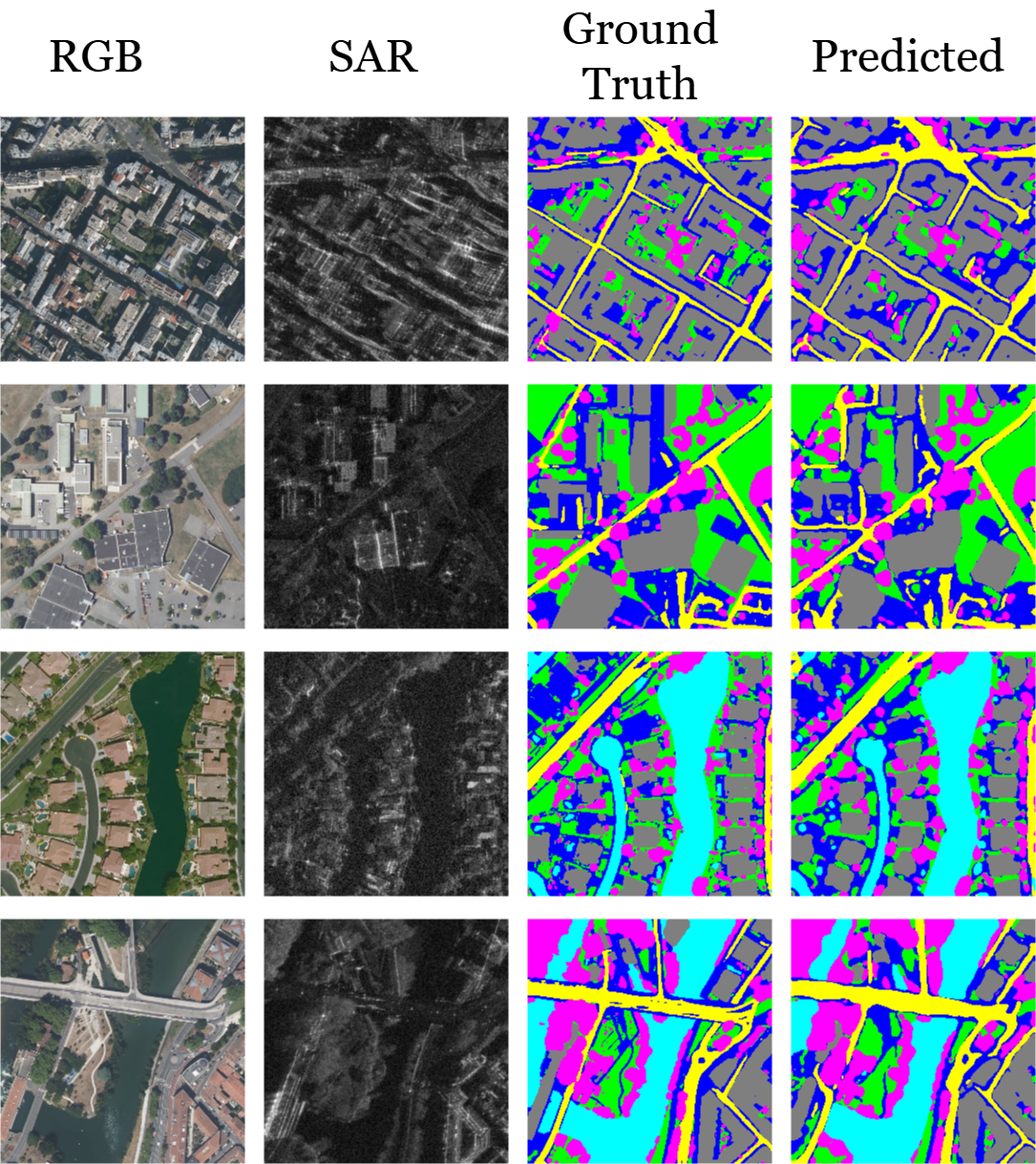} 
    \caption{
        Visualization of our model prediction in the OpenEarthMap-SAR dataset.
    }
    \label{fig:OEM_vis}
\end{figure}

\begin{figure*}[ht]
    \centering
    \includegraphics[scale=0.4]{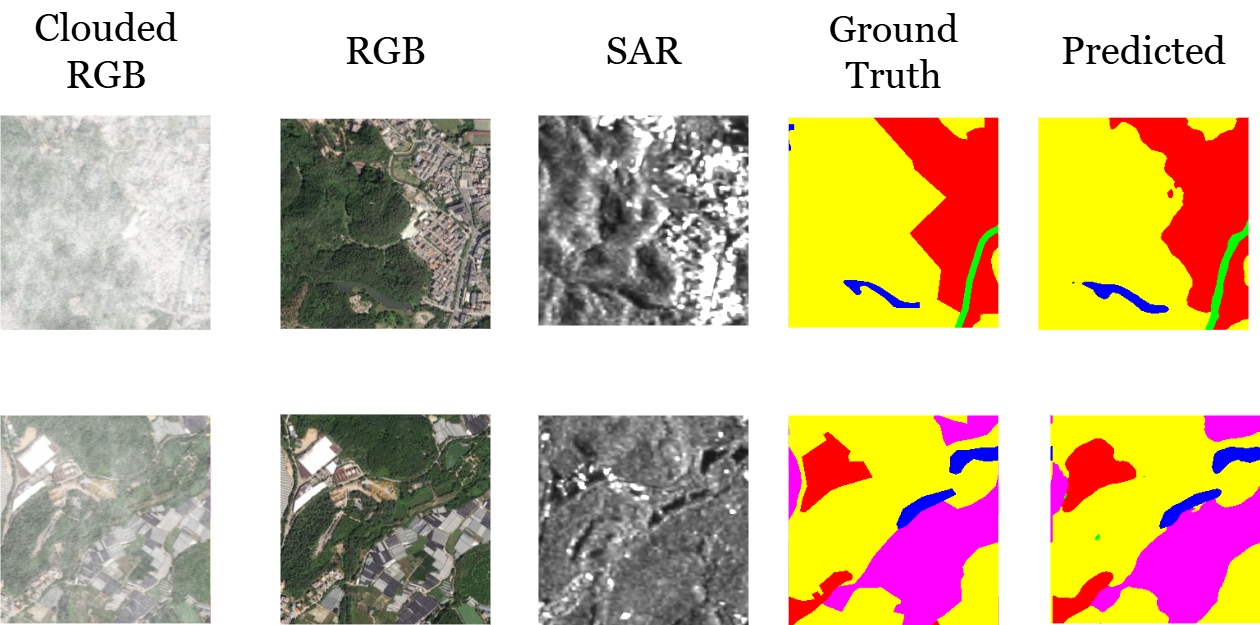} 
    \caption{
        Segmentation result visualization on the PIE-RGB-SAR test set with clouded RGB images. Columns shows cloud-contaminated RGB, reference clear RGB, SAR input, ground truth, and prediction respectively.
    }
    \label{fig:PIE_vis}
\end{figure*}

To further evaluate the effectiveness of our cross-modality segmentation approach, we present visualizations of representative model predictions on the OpenEarthMap-SAR and PIE-RGB-SAR test sets. Figures~\ref{fig:OEM_vis} and~\ref{fig:PIE_vis} provide qualitative comparisons between the RGB, SAR, ground truth, and predicted segmentation maps.

Figure~\ref{fig:OEM_vis} displays segmentation examples from the OpenEarthMap-SAR dataset. Each row represents the RGB input, the SAR input, ground truth, and predicted segmentation. Our method consistently produces spatially coherent predictions that closely match the reference annotations across a wide variety of urban and rural landscapes. Boundaries between classes such as water, built-up areas, and vegetation are well preserved, demonstrating the model's ability to leverage both spectral and structural cues from the multimodal data. Minor errors typically occur at class boundaries or in visually ambiguous regions. Overall, the method achieves robust and visually appealing segmentation.

Figure~\ref{fig:PIE_vis} shows the results on the challenging PIE-RGB-SAR dataset, where some of the heavy cloud coverage from the sample. In the figure, we present the cloud-contaminated RGB input, the corresponding clear reference RGB (for visual context), the SAR input, ground-truth segmentation, and our model's prediction. These results highlight the core advantage of our fusion architecture. Even under severe cloud occlusion, where RGB information is obscured or unable to maintain an accurate semantic delineation of land cover classes. In particular, critical classes such as roads and water bodies remain sharply segmented despite the absence of clear optical cues, underscoring the value of SAR for filling gaps left by clouded optical bands.

\subsection{Language-Based Reasoning}
\label{reasonning}
To enhance the interpretability of our multimodal segmentation results, we employ a metric-driven explainability pipeline using a small language model (SLM) using Phi-3 \cite{Abdin2024Phi3TR}. This process generates concise reasoning for each prediction with quantitative evidence derived from the model outputs.

For every test sample, a comprehensive set of metrics was extracted after inference. These included standard segmentation metrics, such as the mean intersection-over-Union (mIoU), the Dice coefficient, overall accuracy (OA) and Cohen’s kappa. Besides, we calculated a range of indicators tailored to multi-modal fusion. Specifically, for each class, we calculated the predicted and ground-truth coverage percentages, class-wise absolute and signed errors, and detection rates. In addition, we derived fusion-specific metrics such as
fusion quality, RGB and SAR dominance scores, complementarity score, prediction precision, and systematic bias. These metrics provided a holistic summary of the performance of the model, taking into account not only the overall accuracy but also detailed insights into cross-modal integration and the relative contribution of each modality. 

\begin{figure*}[ht!]
    \centering
    \includegraphics[scale=0.4]{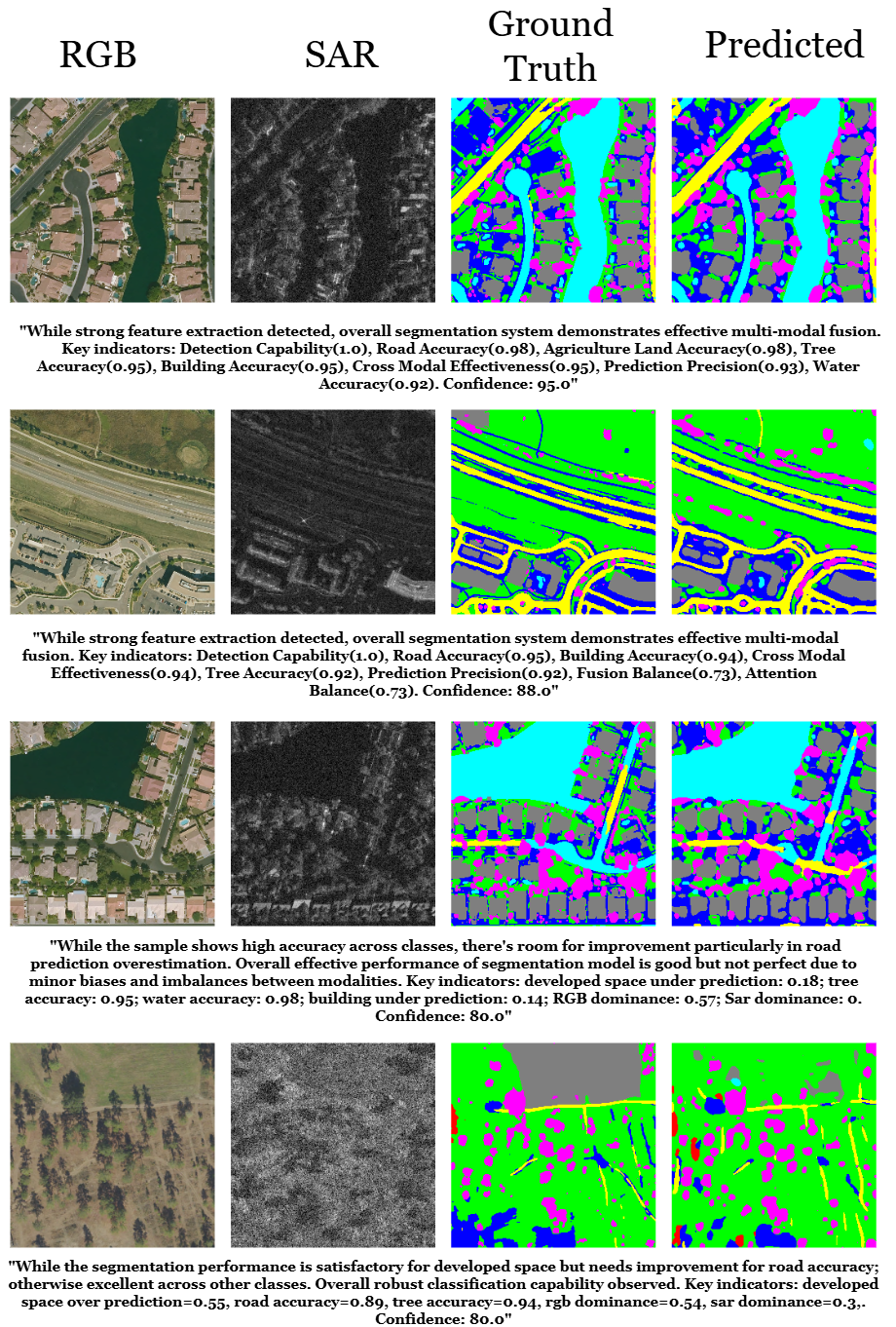}
    \caption{
        Visualization of per-sample reasoning on OpenEarthMap-SAR. Each row illustrates a distinct scene with input RGB and SAR imagery, ground truth, predicted segmentation, and Phi-3 output.
    }
    \label{fig:reasoning}
\end{figure*}

All calculated metrics were organized and encoded in a structured prompt tailored for use by the SLM. This prompt outlined the quantitative framework of the current prediction, featuring summary statistics for each metric and contextual notes on specific patterns related to modality or class. The SLM then processed this prompt to formulate a concise scenario-aware reasoning analysis for each sample. Instead of simply rephrasing metrics, the SLM was instructed to act as a satellite imagery domain expert. The purpose of this prompt is to emphasize the significant strengths and weaknesses in the outcome of segmentation, identifying systematically misclassified classes, and examining the contribution of each modality. The reasoning analysis explicitly assessed whether the model performance was primarily influenced by optical features, SAR features, or their integration, and identified any evidence of complementarity or redundancy between the modalities. Figure \ref{fig:reasoning} shows some examples of reasoning output in conjunction with visualization. By integrating both conventional segmentation scores and custom fusion-aware metrics into the reasoning pipeline, our approach enables a much more transparent and actionable interpretation of model behavior. The framework supports transparent reporting, facilitates root-cause analysis of errors, and provides actionable feedback for model refinement. In practice, this means that beyond reporting a single accuracy number, we can provide context-specific insights on why the model succeeded or failed, which classes or scenes posed challenges, and how future models might be improved through targeted architecture or data augmentation strategies.

\section{Discussion}
\label{discussion}
This work addresses a critical gap in multimodal land cover segmentation: the need for models that can handle cross-modal inconsistencies, class imbalance, and real-world variability while remaining interpretable. Prior studies have made notable advances in fusion techniques, such as local or global attention \cite{ren2024swintfnet}\cite{wang2024mffnet}\cite{quan2024learning} and multiscale feature extraction \cite{liu2025oshfnet}. However, many of these approaches are evaluated under controlled conditions—typically on datasets like WHU-OPT-SAR—where the data are well-aligned, class imbalance is moderate, and atmospheric conditions are ideal.

In contrast, the OpenEarthMap-SAR and PIE-RGB-SAR datasets used in our experiments introduce more realistic and challenging scenarios. These include severe class imbalance (e.g., Bareland and Water classes), cloud-covered optical imagery, and highly diverse geographic regions. Under such conditions, prior models like ASANet \cite{zhang2024asanet}, MFFNet \cite{wang2024mffnet}, and GMFNet \cite{quan2024learning} show diminished performance, particularly in minority classes or visually degraded scenes. CLAIRE is explicitly designed to overcome these issues through its key architectural and methodological contributions.

First, the RIFT loss combines the strengths of Weighted Focal and Tversky Losses, enabling robust learning on underrepresented classes which is a limitation commonly overlooked in earlier models. As shown in Table \ref{tab:ablation_components}, CLAIRE achieves higher rare-class IoUs than alternatives using standard cross-entropy or even class-weighted loss formulations. This is critical for practical applications such as flood mapping or habitat monitoring, where minority land cover types carry high ecological or infrastructural importance.

Second, CLAIRE’s Cross-Modality Attention Fusion (CMAF) module is more adaptive than traditional fusion strategies. Prior works often perform early or late fusion without explicitly modeling inter-modality relevance at multiple scales. CMAF applies spatial and channel-wise attention, gating, and modality-aware enhancement to selectively integrate optical and SAR features. This dynamic fusion is especially advantageous under real-world conditions such as partial cloud cover (as in PIE-RGB-SAR), where reliance on a single modality (optical or SAR) leads to performance degradation. As shown in Table \ref{tab:pie_rgb_sar_ablation} and Figure \ref{fig:OEM_vis}, CLAIRE maintains high segmentation quality even when one modality is partially compromised, outperforming previous methods that suffer from over-reliance on optical features.

Third, the dual encoder architecture enables CLAIRE to extract modality-specific features before fusion, a design choice that contrasts with shared-encoder approaches (e.g., FuseNet \cite{hazirbas2016fusenet}, SA-Gate \cite{chen2020bi}. Shared encoders risk losing unique spatial or spectral cues, particularly when input modalities are heterogeneous and structurally dissimilar, as with SAR and RGB imagery. Our results show that removing the dual encoder leads to significant drops in mIoU and rare-class accuracy.

Furthermore, CLAIRE extends beyond traditional segmentation by introducing a metric-driven interpretability framework using the Phi-3 small language model. While most prior studies stop at producing segmentation maps, CLAIRE explains each prediction by analyzing fusion quality, class-wise errors, and modality dominance. This interpretability is crucial for real-life deployment in environmental policy, disaster response, and geospatial planning, where stakeholders require not only high accuracy but also clear justifications for model decisions \cite{Abdin2024Phi3TR} \cite{lai2024lisa}.

Importantly, CLAIRE does not depend on assumptions that are often implicit in prior work, such as perfect image alignment or ideal atmospheric visibility. While our model currently assumes aligned data during training, it performs well even in visually degraded settings, and can be extended to include registration correction modules. Unlike methods optimized only for curated benchmarks, CLAIRE is trained and tested on datasets with real-world variability, improving its transferability to operational systems used by government agencies, NGOs, or urban planners.
Nevertheless, several limitations remain in CLAIRE. The model’s dual-stream design and attention-heavy fusion module introduce increased computational overhead, which may limit its applicability in edge computing or onboard satellite systems. RIFT’s fixed hyperparameters, while effective, may require tuning for datasets with unseen distributions. In addition, CLAIRE’s interpretability module relies on pre-engineered prompts; without interactive user feedback or learned reasoning capabilities, it may miss edge cases or produce generic explanations.

To address these limitations, future work could explore model compression or lightweight distillation to reduce CLAIRE’s resource footprint \cite{wang2022ofa}. Incorporating automatic hyperparameter tuning or dynamic weighting for RIFT loss could reduce reliance on manual configuration. Moreover, adding spatial alignment mechanisms (e.g., deformable convolutions or feature-domain correlation) could improve robustness when inputs are imperfectly registered \cite{liu2025oshfnet}. Finally, joint training with small language models or reinforcement learning-based explanation agents could make the reasoning module more adaptive and user-driven \cite{chen2022pix2seq}.

\section{Conclusion}
\label{conclusion}
In this study, we introduced CLAIRE, a novel multimodal land cover segmentation framework that integrates cross-modal attention fusion, class imbalance handling, and post-hoc reasoning using a Small Language Model. Our method effectively combines the strengths of optical and SAR imagery while explicitly addressing challenges related to modality misalignment, data imbalance, and interpretability. Through extensive evaluations on three benchmark datasets, CLAIRE consistently outperformed existing methods in terms of mIoU, OA, and Kappa metrics.

Beyond segmentation accuracy, CLAIRE distinguishes itself by incorporating a language-based reasoning module, which offers transparent explanations for predictions, a first in this domain. This feature addresses the increasing demand for explainable AI in remote sensing, especially for high-stakes applications like urban planning and disaster monitoring.
While the results are promising, CLAIRE has limitations. The reasoning module currently relies on structured prompts derived from model outputs, which may limit the depth and flexibility of explanations. Additionally, the framework has not yet been tested on highly dynamic or real-time datasets, which could affect generalizability. Future work will focus on enhancing the reasoning capability using more advanced prompting techniques or vision-language pretraining, and extending CLAIRE’s application to streaming or temporally evolving remote sensing data.

Overall, CLAIRE advances the state of the art by coupling segmentation performance with post-hoc interpretability. Its modular design and strong empirical results make it a viable foundation for future research into transparent and trustworthy multimodal geospatial analysis.

\section*{Declarations}
\textbf{Conflict of Interests:} On behalf of all authors, the corresponding author states that there is no conflict of interest.\\
\textbf{Funding:} No external funding is available for this research.\\
\textbf{Data Availability Statement:} This study conducts all the experiments on 3 publicly available datasets: OpenEarthMap-SAR \cite{xia2023openearthmap} WHU-OPT-SAR \cite{li2022mcanet} and PIE-RGB-SAR \cite{zhang2024asanet}.\\
\textbf{Ethics Approval and Consent to Participate}. Not applicable \\
\textbf{Informed Consents:} All the authors have read the manuscript fully and contributed to this research. The authors also provide consent for peer review and publication of this study.\\
\textbf{CRediT authorship contribution statement:}

\textbf{Debopom Sutradhar}: Conceptualization, Methodology, Writing – original draft; 
\textbf{Arefin Ittesafun Abian}: Resources, Literature review, Writing – original draft; 
\textbf{Mohaimenul Azam Khan Raiaan}: Conceptualization, Writing – original draft, Validation, Formal analysis, Writing – review \& editing; 
\textbf{Reem E. Mohamed}: Validation, Formal analysis, Writing – original draft, Writing – review \& editing; 
\textbf{Sheikh Izzal Azid}: Validation, Formal analysis, Writing – review; 
\textbf{Sami Azam}: Validation, Formal analysis, Writing – review \& editing, Supervision, Project administration.


\end{document}